\pdfoutput=1
\documentclass[11pt]{article}

\usepackage[acceptedwithA]{requirements/acl}
\usepackage{times, latexsym}
\usepackage[T1]{fontenc}
\usepackage[utf8]{inputenc}
\usepackage{microtype}
\usepackage{graphicx}
\usepackage{algorithm}
\usepackage{algpseudocode}
\usepackage{multirow}
\usepackage{amsmath, amsfonts, amssymb}
\usepackage{booktabs}
\usepackage{xspace}
\usepackage{subcaption}
\usepackage[most]{tcolorbox}
\usepackage{siunitx}
\usepackage{setspace}

\usepackage{tikz}
\usetikzlibrary{topaths}
\tikzstyle{every picture}+=[remember picture,inner xsep=0,inner ysep=0.25ex]

\newcommand{\numb}[1]{\num[group-separator={,}]{#1}}

\definecolor{boxcol}{HTML}{DBECFF}
\definecolor{wrongcol}{HTML}{FFE0BD}
\definecolor{corgreen}{HTML}{367000}
\definecolor{highlight}{HTML}{699BD2}
\definecolor{answercol}{HTML}{69D3B5}
\definecolor{answercol}{HTML}{69D3B5}
\definecolor{qcolor}{HTML}{8569D3}
\definecolor{randcol}{HTML}{FFCA8F}

\newcommand{\hlbox}[1]{\colorbox{highlight}{#1}}
\newcommand{\hllbox}[1]{\colorbox{highlight}{\ensuremath{#1}}}

\usepackage{fancyhdr}
\fancypagestyle{firstpage}{
    \fancyhf{}
    \fancyfoot[R]{\textcolor{red}{\emph{Pre-print}}; Accepted to \href{https://transacl.org/index.php/tacl}{\textbf{TACL}} (01/12/2026).}

}

\title{KisMATH: Do LLMs Have Knowledge of Implicit Structures in Mathematical Reasoning?}

\author{
  \textbf{Soumadeep Saha\textsuperscript{$\dagger$}*},
  \textbf{Akshay Chaturvedi\textsuperscript{$\ddagger$}},
  \textbf{Saptarshi Saha\textsuperscript{$\dagger$}},
  \textbf{Utpal Garain\textsuperscript{$\dagger$}},
  \textbf{Nicholas Asher\textsuperscript{*}} \\\\
  \textsuperscript{$\dagger$}ISI Kolkata, India; \textsuperscript{*}IRIT; \textsuperscript{$\ddagger$}LINAGORA Labs; Toulouse, France \\
  \small{\textbf{Correspondence:} \href{mailto:soumadeep.saha97@gmail.com}{soumadeep.saha97@gmail.com}}
}

\begin{document}
\thispagestyle{firstpage}
\maketitle

% ----------------------- CHAPTERS ---------------------------------------------
\begin{abstract}
    Chain-of-thought (CoT) traces have been shown to improve performance of large
    language models on a plethora of reasoning tasks, yet there is no consensus
    on the mechanism by which this boost is achieved. To shed more light on
    this, we introduce Causal CoT Graphs (CCGraphs), which are directed acyclic
    graphs automatically extracted from reasoning traces that model fine-grained
    causal dependencies in language-model outputs. A collection of \numb{1671}
    mathematical reasoning problems from MATH500, GSM8K, and AIME, together with
    their associated CCGraphs, has been compiled into our
    dataset---\textbf{KisMATH}. Our detailed empirical analysis with \numb{15}
    open-weight LLMs shows that (i) reasoning nodes in the CCGraphs are causal
    contributors to the final answer, which we argue is constitutive of
    reasoning; and (ii) LLMs emphasize the reasoning paths captured by the
    CCGraphs, indicating that the models internally realize structures similar
    to our graphs. KisMATH enables controlled, graph-aligned interventions and
    opens avenues for further investigation into the role of CoT in LLM
    reasoning.
\end{abstract}

\section{Introduction}
\label{sec:intro}

\begin{figure*}[t!]
    \centering
    \begin{subfigure}[t]{0.49\textwidth}
        \begin{center}
            \begin{tcolorbox}[width=\linewidth,colback=wrongcol,colframe=black, equal height group=A] \footnotesize{
                \textbf{Question: } Mike plays ping pong for 40 minutes. In the first 20
                minutes, he scores
                \tikz[baseline = (n1.base)] \node (n1){\colorbox{qcolor}{$4$}};
                points. In the second 20 minutes, he scores
                \tikz[baseline = (n2.base)] \node (n2){\colorbox{qcolor}{$25\%$}};
                more points. How many total points did he score?\\\\
                \textbf{Reasoning:} \\
                - Points scored in the first 20 minutes: 
                \tikz[baseline = (n3.base)] \node (n3){\hlbox{$4$}};\\
                -
                \tikz[baseline = (n4.base)] \node (n4){\hlbox{$25\%$}};
                of 
                \tikz[baseline = (n5.base)] \node (n5){\hlbox{$4$}};
                points is
                \tikz[baseline = (n6.base)] \node (n6){\hlbox{$0.25 \times 4$}};
                $\,=\,$
                \tikz[baseline = (n7.base)] \node (n7){\hlbox{$1$}};
                point.\\
                - Points scored in the second 20 minutes:\\\\
                \tikz[baseline = (n8.base)] \node (n8){\hlbox{$4 + 1$}};
                $\,=\,$
                \tikz[baseline = (n9.base)] \node (n9){\hlbox{$5$}};.\\
                - Total points scored:
                \tikz[baseline = (n10.base)] \node (n10){\hlbox{$4 + 5$}};
                $\,=\,$
                \tikz[baseline = (n11.base)] \node (n11){\hlbox{$9$}};.\\\\
                \textbf{Answer: } 
                \tikz[baseline = (n12.base)] \node (n12){\colorbox{answercol}{$9$}};.
            }
            \begin{tikzpicture}[overlay]
                \draw[-latex] (n1.south) to[out=270, in=90] (n3.north);
                \draw[-latex] (n2.south) to[out=270, in=90] (n4.north);
                \draw[-latex, gray] (n3.west) to[out=180, in=90] (n5.north);
                \draw[-latex, gray] (n4.north) to[out=45, in=135] (n6.north);
                \draw[-latex, gray] (n5.east) to[out=45, in=135] (n6.west);
                \draw[-latex, gray] (n6.east) to[out=45, in=135] (n7.west);
                \draw[-latex, gray] (n7.south) to[out=225, in=90] (n8.north);
                \draw[-latex, gray] (n5.south) to[out=270, in=90] (n8.north);
                \draw[-latex, gray] (n8.east) to[out=45, in=135] (n9.west);
                \draw[-latex, gray] (n9.east) to[out=0, in=90] (n10.north);
                \draw[-latex, gray] (n5.south) to[out=270, in=90] (n10.north);
                \draw[-latex, gray] (n10.east) to[out=45, in=135] (n11.west);
                \draw[-latex] (n11.south) to[out=270, in=0] (n12.east);
                \draw[-latex] (n10.south) to[out=270, in=0] (n12.east);
            \end{tikzpicture}
            \end{tcolorbox}
        \end{center}
    \end{subfigure}
    \begin{subfigure}[t]{0.49\textwidth}
        \begin{center}
            \begin{tcolorbox}[width=\linewidth,colback=wrongcol,colframe=black, equal height group=A] \footnotesize{
                \textbf{Question: } Mike plays ping pong for 40 minutes. In the first 20
                minutes, he scores
                \tikz[baseline = (n1.base)] \node (n1){\colorbox{qcolor}{$4$}};
                points. In the second 20 minutes, he scores $25\%$ more points.
                How many total points did he score?\\\\
                \textbf{Reasoning:} \\
                - Points scored in the
                \tikz[baseline = (r3.base)] \node (r3){\colorbox{randcol}{first}};
                20 minutes: 
                \tikz[baseline = (n3.base)] \node (n3){\hlbox{$4$}};\\
                - $25\%$ of 
                \tikz[baseline = (n5.base)] \node (n5){\hlbox{$4$}};
                points is $0.25 \times 4 = 1$ point.\\
                - Points scored
                \tikz[baseline = (r4.base)] \node (r4){\colorbox{randcol}{in}};
                the second 20 minutes: $4 + 1 = 5$.\\
                -
                \tikz[baseline = (r5.base)] \node (r5){\colorbox{randcol}{Total}};
                points scored:
                \tikz[baseline = (n10.base)] \node (n10){\hlbox{$4 + 5$}};
                $=9$.\\\\
                \textbf{Answer: } 
                \tikz[baseline = (n12.base)] \node (n12){\colorbox{answercol}{$9$}};.
            }
            \begin{tikzpicture}[overlay]
                \draw[-latex] (n1.south) to[out=270, in=90] (n3.north);
                \draw[-latex] (n3.south) to[out=270, in=90] (n5.north);
                \draw[-latex] (n5.south) to[out=270, in=180] (n10.west);
                \draw[-latex] (n10.south) to[out=270,in=0] (n12.east);

                \draw[-latex, dashed, gray] (n1.south) to[out=270, in=90] (r3.north);
                \draw[-latex, dashed, gray] (r3.south) to[out=270, in=90] (r4.north);
                \draw[-latex, dashed, gray] (r4.south) to[out=270, in=90] (r5.north);
                \draw[-latex, dashed, gray] (r5.south) to[out=270, in=90] (n12.north);
            \end{tikzpicture}
            \end{tcolorbox}
        \end{center}
    \end{subfigure}
    \caption{\textbf{Example of extracted causal graph and paths.}
    \emph{(Left)} An example of a (simplified) CoT causal graph (\textbf{CCGraph})
    extracted from the GSM8K dataset. Reasoning nodes are highlighted in blue,
    edges are in gray. \emph{(Right)} An \textbf{R path} (see Eq. \ref{eq:rpath}),
    i.e., a simple path from question to answer (solid line) and a random path
    (dashed line).
    }
    \label{fig:example_annotation}
\end{figure*}

\emph{Chain-of-thought} prompting \citep{wei2022chain} and its subsequent
variants \citep{wang-etal-2023-plan, zhang2023automatic} have been shown to be
effective at eliciting improved performance from large language models (LLMs) at
reasoning-oriented tasks such as mathematics and programming \citep{tacl0, tacl1}.
More recently, large-scale reinforcement-learning (RL) post-training
\citep{deepseek-r1} has demonstrated further performance improvements and
resulted in a class of models referred to as ``large reasoning models'' (LRMs).
Beyond improved performance indicators, these models generate extended \emph{CoT}
rationales (often termed long \emph{CoT}, reasoning traces, \emph{CoT} rollouts,
or derivational traces) that contain a simulacrum of capabilities such as
self-verification, backtracking, and reflection \citep{openai2024openaio1card, deepseek-r1}.

However, there is no consensus on \emph{how CoT works}, with opinions largely
split into two prominent camps. The first argues that \emph{CoT} improves
performance by decomposing complex problems into manageable sub-tasks, solving
these sub-tasks, and recombining the partial solutions (with potential
backtracking or consistency checks) to arrive at an answer
\citep{openai2024openaio1card, deepseek-r1}. The second camp contends that
\emph{CoT} assists with some form of ``approximate retrieval'' from latent
knowledge \citep{Kambhampati_2024, Kambhampati_2025}; and is largely insensitive
to perturbations of (i) in-context examples \citep{wang-etal-2023-towards}, (ii)
the training data \citep{li2025llmseasilylearnreason, stechly2025semantics}, or
(iii) the post-training RL reward function \citep{shao2025spuriousrewards}.

In this paper, we shed more light on this pressing question; to wit, \emph{our
contributions are} as follows:
\begin{itemize}
    \item \emph{We devise an algorithm to extract causal graphs from
    LLM-generated derivational traces}. The nodes of these graphs are
    mathematical expressions and its edges capture fine-grained causal
    dependencies between them. The resulting ``\emph{Causal CoT graph}''
    (\textbf{CCGraph}) expresses the latent structure that links a question to
    its answer via intermediate computations (see Figure \ref{fig:example_annotation}).
    \item Using this algorithm, we construct \textbf{KisMATH} (\textbf{K}nowledge
    of \textbf{i}mplicit \textbf{s}tructures in \textbf{Math})---a dataset
    consisting of \numb{1671} mathematics problems drawn from GSM8K
    \citep{gsm8k}, MATH500 \citep{math500} and AIME \citep{aime}, paired with
    its LLM generated solution and \textbf{CCGraph}.
    \item \emph{We analyze a wide array (15) of open-weight LLMs ranging from
    1B--70B parameters and find that: (i) mathematical expressions in reasoning
    traces are effective mediators between a question and the final answer, which we
    argue is a constituent condition for reasoning, and (ii) these models
    demonstrate properties suggesting that they implicitly realize structures
    similar to the proposed causal CoT graph.}
\end{itemize}

\textbf{KisMATH} elucidates the fine-grained causal structure underlying the
reasoning trace, and, unlike previous studies that relied on stochastic
perturbations, enables the exploration of controlled, graph-based interventions
in mathematical reasoning. The dataset, along with all supporting artifacts, is
available at
\href{https://espressovi.github.io/KisMATH}{espressovi.github.io/KisMATH}.

\section{Background}
\label{sec:background}

Despite the remarkable performance gains observed with \emph{CoT}, a mounting
body of evidence points to the fact that models don't ``reason'' in accordance
with a conventional understanding of the term. For example,
\citet{li2025llmseasilylearnreason} demonstrated that distilling a model on
reasoning traces in which up to $50\%$ of the numbers are randomly replaced does
not significantly affect performance---a finding corroborated by
\citet{stechly2025semantics} in the context of planning problems. Perturbations
in \emph{in-context} demonstrations have also been demonstrated to have minimal
effect \citep{wang-etal-2023-towards}, and \citet{shao2025spuriousrewards} have
reported performance improvements with spurious rewards during the post-training
RL step. 

Furthermore, several conventional expectations from reasoning, such as
intermediate reasoning steps being reliable causal mediators for the final
answer \citep{paul-etal-2024-making}, or the generated rationales providing
faithful explanations for the final answer \citep{lanham2023measuringfaithfu,
lyu-etal-2023-faithful}, have been demonstrated not to be true for LLM
reasoning. \citet{barez-chain-2025}, in their survey, highlighted that \emph{CoT}
traces do not provide faithful explanations for the final answers. Significant
performance drops have also been observed with increasing ``problem size''
\citep{stechly2024chain}, which is contrary to expectations from reasoning
models.

In this work, we first extract the causal structure from an LLM-generated
reasoning trace for a variety of mathematical reasoning tasks, and ask whether
this structure---implicit in any reasoning response---has any \emph{special
relevance} for LLMs. As opposed to randomized interventions
\citep{li2025llmseasilylearnreason, wang-etal-2023-towards}, interventions
aligned with the implicit structure of reasoning are more informative.
\citet{tan-2023-causal} noted that interventions at the node level in
mathematical reasoning traces often led to ``self-correcting behavior'' from
LLMs. In their experiments, intervening on a node in the causal graph introduced
calculation errors into the reasoning trace, which could be determined from
surrounding context, and corrected by the LLM. Owing to the complex,
interwoven nature of multi-step reasoning, interventions which do not take this
implicit structure into account offer weaker evidence towards the role of
\emph{CoT} traces on the final outcome.

Previous work in attempting to capture structure in LLM reasoning, such as works
by \citet{tan-2023-causal, lee2025reasoningflow, bogdan2025thoughtanchors}, has
been limited in scale. \citet{tan-2023-causal} manually annotated causal graphs
for $27$ GSM8K examples, and \citet{lee2025reasoningflow,
bogdan2025thoughtanchors} annotated $30, 10$ reasoning traces respectively.
The annotations by \citet{lee2025reasoningflow} and
\citet{bogdan2025thoughtanchors} are much richer, consisting of various edges
corresponding to computation, planning, backtracking, reflection, etc., but the
general problem of decomposing a \emph{CoT} trace into sound atomic inferences
and capturing their relationships continues to pose a challenge
\cite{jacovi-etal-2024-chain, golovneva2023roscoe}.
\citet{mukherjee2025premiseaugmented} introduced an LLM-based framework to
identify the premises involved in an inference step to help guide the reasoning
process, however, such a framework is inherently limited by the capabilities of
the employed LLM. \citet{bogdan2025thoughtanchors} proposed three techniques for
annotating relationships in reasoning traces---rollout sampling, attention
aggregation, attention suppression---at the sentence level. They noted that the
attention aggregation technique is not a reliable proxy for causality, and the
rollout sampling technique is computationally prohibitive, especially for
extracting more granular relationships. We employed attention suppression for
validating our extracted graphs.

Our approach to extracting \textbf{CCGraphs} is much more scalable; we present
the \textbf{KisMATH} dataset containing $1671$ annotated \emph{CoT} traces for
mathematical reasoning tasks. The \textbf{CCGraphs} capture detailed mathematical
causal relationships (see Figure \ref{fig:example_annotation}) and feature 9--40
intermediate reasoning nodes per problem and 6--10 reasoning hops from the query
to the answer (see Table \ref{tab:dataset-char}).

In previous work, attempts have been made to interpret how LLMs track entities
and their state evolution \citep{li-etal-2021-implicit, fagnou-etal-2024-chain}.
Techniques to incorporate entity relationships \citep{liu-etal-2024-era},
information about premises \citep{mukherjee2025premiseaugmented}, and discourse
structure \citep{sharma2025dimsumdisc}---to aid reasoning performance---have
also been explored. In contrast, our work attempts to probe LLMs regarding their
``\emph{awareness}'' of relationships between mathematical entities and the role
the latter play in arriving at a solution to mathematical reasoning problems.

\section{Dataset: KisMATH}
\label{sec:method}

In this section we outline the process of curating the \textbf{KisMATH} dataset.
Mathematical-reasoning problems are sourced from the popular mathematics
benchmark datasets: GSM8K \citep{gsm8k}, MATH500 \citep{math500}, and the AIME
dataset \citep{aime}. Although some of these datasets provide accompanying
solutions, for a more general treatment we generated \emph{CoT} rollouts from
OpenAI \texttt{o3} for all selected problems. Given the question, the \texttt{o3}
generated reasoning response, and the answer\footnote{Only responses with
\textbf{correct answers} were selected.}, we first extract and parse all
mathematical expressions present in each \emph{CoT} trace (with a symbolic
parser - \texttt{SymPy}). Employing this list of parsed expressions, we
construct a graph---termed the \emph{Causal CoT Graph} (\textbf{CCGraph})---by
starting from the node corresponding to the answer, expanding to nodes that
match the answer, and from each node recursively expanding until we reach nodes
which are part of the question. All edges are then reversed to arrive at the
final \textbf{CCGraph}. This process results in $983$, $384$, and $304$
(question, reasoning, \textbf{CCGraph}, answer) 4-tuples for the GSM8K, MATH500,
and AIME splits, respectively.

\subsection{Causal CoT Graph Construction}
\label{subsec:ccgc}

Given a question $Q$, reasoning trace $R$, and answer $A$ we first extract
mathematical expressions (e.g., numbers, \LaTeX{} formulas) which are spans from
$Q$, $R$, and $A$. We further ensure that all spans are
non-intersecting\footnote{If two extracted spans intersect, their union is used.},
and sort them in order of their start indices (earliest-first), giving
\begin{align*}
    \hat{Q} &= [\hat{q}_{1}, \hat{q}_{2}, \ldots \hat{q}_{(n_Q)}]\\
    \hat{R} &= [\hat{r}_{1}, \hat{r}_{2}, \ldots \hat{r}_{(n_R)}]\\
    \hat{A} &= [\hat{a}] \tag*{(datasets have one answer)}
\end{align*}
where each $\hat{q}_{i}, \hat{r}_{i},$ and $\hat{a}$ is a span in the question,
reasoning, and answer, respectively. With these 3 sorted lists of
non-intersecting spans $\hat{Q}$, $\hat{R}$, $\hat{A}$, we construct a
\textbf{CCGraph} corresponding to each sample according to Algorithm
\ref{alg:graph-construct}.

\begin{algorithm}
\caption{\textbf{: CCGraph} Construction}
\label{alg:graph-construct}
\begin{algorithmic}[1]
    \State Given $\hat{Q}$, $\hat{R}$, and $\hat{A} = [\hat{a}]$.
    \State $G \leftarrow (\{\hat{a}\}, \phi)$ \Comment{Initial \textbf{CCGraph}}
    \State \texttt{context} $\leftarrow \text{concatenate}(\hat{Q}, \hat{R})$
    \State \Call{Expand}{$\hat{a}$, \texttt{context}, $G$}
    \State \Call{Prune}{$G$} \Comment{Nodes with no path to some $\hat{q}$.}
    \State Reverse all edges in $G$. \Comment{Result \textbf{CCGraph}}
    \item[]
    \Procedure{Expand}{$\hat{i}$, \texttt{context}, $G$}
        \If{$|\texttt{context}| \leq |\hat{Q}|$}
            \State \texttt{return} \Comment{$\hat{q}\rightarrow\hat{q}'$ edges skipped.}
        \EndIf
        \For{$\hat{j} \in \text{reversed}(\texttt{context})$}
            \State $p_i, p_j \leftarrow \Call{Parse}{\hat{i}}, \Call{Parse}{\hat{j}}$
            \If{ \Call{match}{$p_i, p_j$} }
                \State Add \emph{node} $\hat{j}$ to graph $G$.
                \State Add \emph{edge} $(\hat{i} \rightarrow \hat{j})$ to $G$.
                \State \Call{Expand}{$\hat{j}$, \texttt{context}[$<\hat{j}$], $G$}
            \EndIf
        \EndFor
    \EndProcedure
\end{algorithmic}
\end{algorithm}

Nodes $\hat{q} \in \hat{Q}$, $\hat{r} \in \hat{R}$, and $\hat{a}$ are referred
to as question nodes, reasoning nodes, and the answer node, respectively. Two
parsed expressions $p_i, p_j$ are said to match if they are exact string matches
or their parse trees share a common node. For example, in Figure
\ref{fig:example_annotation}, the node ``$4$'' matches the node ``$4 + 5$'', as
$4$ contributes to the sum ``$4 + 5$''. The procedure starts searching for
matches of the answer with every other node (\texttt{context}), and whenever a
match is found, a node and edge is added, which becomes another candidate search
query. For any search query, every term that appears before the query
(\texttt{context}[$<\hat{j}$]) is considered for matching, to ensure that the
constructed \textbf{CCGraph} is \emph{directed-acyclic} (DAG). The condition on
line \texttt{8} (Algorithm \ref{alg:graph-construct}) checks if the
\texttt{context} is entirely composed of question nodes, in which case the
search is terminated since we are not interested in studying relationships
between question nodes.

Since we analyze the effect of reasoning nodes on the final answer, given the
constructed graph $G$, we prune it (line \texttt{5}, Algorithm
\ref{alg:graph-construct}) to remove all nodes that do not have a path to at
least one question node.\footnote{All nodes have a path from the answer node by
construction.} This might result in a singleton graph containing only the answer,
and in such scenarios ($\sim\!10\%$ of cases) we \textbf{manually intervene} to
ensure a non-trivial graph exists. The manual edits made to the
\textbf{CCGraphs} strictly fall into one of the following categories: (i)
\LaTeX{} errors such as missing ``\$'', unmatched ``\{'', etc., or (ii) natural
language interruptions, i.e., situations where an equation is made un-parseable
owing to the presence of natural language. For example the expression ``$4+5$ is
$9$'' is replaced with ``$4+5 = 9$'', or ``$5$ is greater than $4$'' is replaced
with $5 > 4$, etc. Such interventions were only made when the absence of such an
edge rendered the \textbf{CCGraph} disconnected. Only $88$ out of the $\sim40K$
parsed vertices and $71$ out of the $\sim300K$ extracted edges required manual
intervention.

For further analysis in this work, \emph{we selected the top-k longest} $Q
\leadsto A$ \emph{paths}, i.e., $k$ \emph{longest unique directed simple paths
that start from any question node and end at the answer node} (via reasoning
nodes) from our \textbf{CCGraphs}.\footnote{$k=5$ for GSM8K, $k=10$ otherwise.}
These are referred to as \textbf{reasoning paths} or \textbf{R paths} in the
following sections. More explicitly, an \textbf{R path} looks like 
\begin{equation}
    [\hat{q}_{\alpha} \rightarrow \hat{r}_{(i_1)} \rightarrow \hat{r}_{(i_2)} \rightarrow \ldots \hat{r}_{(i_{\mu})} \rightarrow \hat{a}]
    \label{eq:rpath}
\end{equation}
where $i_1 < i_2 < \ldots i_{\mu}$. Figure \ref{fig:example_annotation}
\emph{(right)} shows an example of a \textbf{CCGraph} constructed following
Algorithm \ref{alg:graph-construct} and a selected \textbf{R path} \emph{(left)}.
Figure \ref{fig:path_examples2} and Figure \ref{fig:path_examples} (Appendix
\ref{sec:appendix-path-examp}) contain further examples of selected \textbf{R
paths} from our dataset.\footnote{Further examples are available at
\href{https://espressovi.github.io/KisMATH}{espressovi.github.io/KisMATH}.}

\subsection{Experimental Details}

The reasoning problems used in the study are sourced from the GSM8K
\citep{gsm8k}, MATH500 \citep{math500} and AIME (1983 - 2024) \citep{aime}
datasets. GSM8K is a collection $\sim\!7500$ arithmetic word problems, whereas
MATH500 and AIME contain \numb{500} and \numb{993} Olympiad-style,
pre-calculus-level mathematics problems, respectively, drawn from several domains such as combinatorics, geometry and algebra (examples in Figure
\ref{fig:example_dataset}). In addition to questions and ground-truth answers,
MATH500 and GSM8K also include solutions to the problems; however, for a general
source-independent treatment, \emph{these solutions are not used}. Geometry
problems or problems featuring diagrams in the MATH500 and AIME datasets were
filtered out, as they present additional challenges for mathematical expression
parsing. We chose \numb{1000}, \numb{389}, and \numb{350} samples from GSM8K,
MATH500 and AIME, respectively. Table \ref{tab:dataset-char} lists some
statistics for the \textbf{KisMATH} dataset, and \emph{further experimental
details are presented in Appendix} \ref{sec:appendix-experiments}.

\begin{table}
    \begin{center}
        \footnotesize{
        \begin{tabular}[c]{p{3em}|l c r}
            \toprule
            Attribute         & \textbf{GSM8K}  & \textbf{MATH500}  & \textbf{AIME}     \\
            (avg.)            & (983)           & (384)             & (304)             \\
            \midrule
            $|V|$             & 14.2 $\pm$ 4.6  & 28.7  $\pm$ 17.9  & 50.6 $\pm$ 20.9   \\
            $|E|$             & 40.8 $\pm$ 30.4 & 260.8 $\pm$ 321.2 & 566.6$\pm$ 487.7  \\
            $|\hat{Q}|$       & 3.9  $\pm$ 1.4  & 7.0   $\pm$ 4.7   & 11.1 $\pm$ 6.6    \\
            $|\hat{R}|$       & 9.3  $\pm$ 3.7  & 20.7  $\pm$ 14.8  & 38.5 $\pm$ 18.2   \\
            $\text{len}(r)$   & 6.4  $\pm$ 1.8  & 8.7   $\pm$ 2.1   & 10.9 $\pm$ 1.6    \\
            \bottomrule
        \end{tabular}}
    \end{center}
    \caption{\textbf{Statistics of various splits of the KisMATH dataset.}
    For each problem we construct \textbf{CCGraph} $G_i = (V_i, E_i)$.
    $|\hat{Q}_i|, |\hat{R}_i|$ is the number of parsed graph nodes that are part
    of the question and reasoning, respectively. $\text{len}(r)$ refers to the
    lengths of the paths chosen for analysis (\textbf{R paths}) in this study.
    }
    \label{tab:dataset-char}
\end{table}

To generate \emph{CoT} traces, OpenAI \texttt{o3-2025-04-16} \citep{o3} was used
with hand-crafted \emph{5-shot CoT prompts} specific to each split. Additional
instructions were provided to improve reasoning structure, \LaTeX{} formatting,
etc., to assist with subsequent steps (example prompt in Appendix
\ref{sec:appendix-prompts}, Figure \ref{fig:prompt}). All generation settings
were at their default values, and the \texttt{reasoning\_effort} parameter was
set to \texttt{medium}.

\textbf{Fifteen} open-weight LLMs were analyzed in the study, ranging from the
1B to 70B parameter scale. These are - Gemma 3 1B, 12B, 27B \citep{gemma3}; Qwen
3 1.7B, 8B, 32B \citep{qwen3}; DeepSeek (DS) R1 1.5B, 8B, 32B, 70B
\citep{deepseek-r1}; Llama 3.1 8B, Llama 3.3 70B \citep{llama3, llama31,
llama33}, Qwen 2.5 7B, Qwen 2.5 7B Math \citep{qwen}, and DeepSeek R1 0528 8B
\citep{deepseeknew}. Instruction-tuned variants of models were used whenever
available, and all models were prompted with \emph{5-shot CoT prompts}
(identical to \texttt{o3}) for all experiments. Temperature was set to $T=1$,
and all other generation parameters were disabled. All models were acquired from
\href{https://huggingface.co}{HuggingFace} and implemented in
\href{https://pytorch.org/}{PyTorch}. Barring \texttt{o3}, which was accessed
through the OpenAI API, and results in Figure \ref{fig:perf-var}, which were
acquired through the \href{https://openrouter.ai/}{OpenRouter} API, all
inference was carried out on $4 \times$ A100 GPUs. The total compute used in the
study is $\sim 3000$ GPU-hours (+\$50 in API usage). 

\begin{figure*}[t!]
    \begin{center}
    \begin{subfigure}[t]{0.49\textwidth}
        \begin{tcolorbox}[width=\linewidth,colback=wrongcol,colframe=black, equal height group=aac] \footnotesize{
            \textbf{Question}: Let $a$ and $b$ be positive real numbers such that $ab^2 = \colorbox{qcolor}{5}$.  Find the minimum value of $a^4 + b^8$.\\
            \textbf{{Reasoning}}:\\
            Let's think step by step.  
            - The constraint gives $ab^{2}=5 \implies a=\dfrac{5}{b^{2}}$.  
            - Substitute into the objective: $a^{4}+b^{8}=(\dfrac{5}{b^{2}})^{4}+b^{8}=\dfrac{5^{4}}{b^{8}}+b^{8}$.  
            - Compute $5^{4}=\hllbox{625}$, so the expression is $\dfrac{625}{b^{8}}+b^{8}$.  
            - Let $x=b^{8}$ with $x>0$; then the expression becomes $g(x)=\dfrac{625}{x}+x$.  
            - Differentiate: $\hllbox{g'(x)}=-\dfrac{625}{x^{2}}+1$.  
            - Set $\hllbox{g'(x)}=0$: $1=\hllbox{\dfrac{625}{x^{2}}}\implies x^{2}=625\implies x=25$ (since $x>\hllbox{0}$).  
            - Verify a minimum: $\hllbox{g''(x)}=\dfrac{1250}{x^{3}}>0$, so $x=\hllbox{25}$ indeed gives the minimum.  
            - Hence $b^{8}=\hllbox{25}$ and the minimum value is $g(25)=\dfrac{625}{25}+25=25+25=\hllbox{50}$.\\
            \textbf{Answer}: \colorbox{answercol}{50}
        }\end{tcolorbox}

        \begin{tcolorbox}[width=\linewidth,colback=wrongcol,colframe=black, equal height group=bbb] \footnotesize{
                \textbf{Question}: In a class of 50 students, \colorbox{qcolor}{28} participate in MATHCOUNTS, 21 participate in science club, and 6 students participate in neither. How many students participate in both MATHCOUNTS and science club?\\
            \textbf{{Reasoning}}:\\
            Let's think step by step.
            - Let $\hllbox{|M|}=28$, $|S|=21$, and $|M \cap S|=x$.
            - $6$ students do neither, so $\hllbox{|M \cup S|}=50-6=44$.
            - By inclusion–exclusion, $|M \cup S|=\hllbox{|M|+|S|-|M \cap S|}$.
            - Substitute: $\hllbox{44 }= 28 + 21 - x$.
            - Solve: $x = 28 + 21 - 44 =\hllbox{ 5}$.\\
            \textbf{Answer}: \colorbox{answercol}{5}
        }
        \end{tcolorbox}

        \caption{\textbf{MATH500}}
    \end{subfigure}
    \begin{subfigure}[t]{0.49\textwidth}

        \begin{tcolorbox}[width=\linewidth,colback=wrongcol,colframe=black, equal height group=aac] \footnotesize{
            \textbf{Question}: What is the largest $2$-digit prime factor of the integer $n = \colorbox{qcolor}{\ensuremath{{200\choose 100}}}$ ?\\
            \textbf{{Reasoning}}:\\
            Let's think step by step.  
            - For a prime $p$, the exponent of $p$ in $\binom{200}{100}$ is $e_p=\lfloor\dfrac{200}{p}\rfloor+\lfloor\dfrac{200}{p^{2}}\rfloor-2\lfloor\dfrac{100}{p}\rfloor-2\lfloor\dfrac{100}{p^{2}}\rfloor$.
            - For any two–digit prime $p\ge 11$ we have $p^{2}>100$, so $\lfloor\dfrac{100}{p^{2}}\rfloor=0$.  
            - If $p\ge 17$ then $p^{2}>200$, so $\lfloor\dfrac{200}{p^{2}}\rfloor=0$ as well, giving $e_p=\lfloor\dfrac{200}{p}\rfloor-2\lfloor\dfrac{100}{p}\rfloor$.  
            - For $p\ge 67$ we have $\lfloor\dfrac{200}{p}\rfloor=2$ and $\lfloor\dfrac{100}{p}\rfloor=1$, so $e_p=2-2\cdot1=0$. Thus no prime $p\ge 67$ divides the binomial coefficient.  
            - Check $p=61$: $\lfloor\dfrac{200}{61}\rfloor=3$, $\hllbox{\lfloor\dfrac{100}{61}\rfloor}=1$, so $\hllbox{e_{61}}=3-2\cdot1=1>0$, hence $\hllbox{61}$ divides $\binom{200}{100}$.  
            - Primes between $\hllbox{61}$ and $67$ are $59$ and $53$; both are smaller than $\hllbox{61}$, so $\hllbox{61}$ is currently the largest divisor found.  
            - Since every larger two–digit prime fails and $\hllbox{61}$ succeeds, the largest two–digit prime factor is $\hllbox{61}$.\\
            \textbf{Answer}: \colorbox{answercol}{61}
        }\end{tcolorbox}

        \begin{tcolorbox}[width=\linewidth,colback=wrongcol,colframe=black, equal height group=bbb] \footnotesize{
            \textbf{Question}: Let $f(x)=|x-p|+|x-15|+|\colorbox{qcolor}{x-p-15}|$, where $0 < p < 15$ . Determine the minimum value taken by $f(x)$ for $x$ in the interval $p \leq x\leq15$ .\\
            \textbf{{Reasoning}}:\\
            Let’s think step by step.  
            - For $p<x<15$ we have $x-p>0$ so $|x-p|=x-p$.  
            - Also $x<15$ so $|x-15|=15-x$.  
            - Since $x\le15$ and $p>0$ we have $\hllbox{x-p-15}<0$ so $|x-p-15|=p+15-x$.  
            - Therefore on $p\le x\le15$,  $\hllbox{f(x)}=x-p+15-x+p+15-x$.  
            - Simplify: $f(x)=\hllbox{15+|x-15|}=30-x$ for $p\le x\le\hllbox{15}$.  
            - $30-x$ decreases as $x$ increases, hence its minimum on $[p,\hllbox{15]}$ is at $x=\hllbox{15}$.  
            - Evaluate: $f(15)=30-15=\hllbox{15}$.\\\textbf{Answer}: \colorbox{answercol}{15}
        }\end{tcolorbox}

        \caption{\textbf{AIME}}
    \end{subfigure}
    \end{center}
    \caption{
        Examples of \textbf{R paths} from the MATH500 and AIME splits of the
        \textbf{KisMATH} dataset. Nodes on the \textbf{R path}
        ($\hat{q}_{\alpha} \leadsto \hat{r}_{(i_1)} \leadsto \ldots \leadsto
        \hat{r}_{(i_{\mu})} \leadsto \hat{a}$) are highlighted (see Eq.
        \ref{eq:rpath}).
    }
    \label{fig:path_examples2}
\end{figure*}

\section{A Causal View of Mathematical Reasoning}
\label{sec:causal}

\citet{paul-etal-2024-making} formulated a causal view of \emph{CoT}-aided
reasoning, wherein they framed the reasoning process as a causal graph (a
probabilistic graphical model describing node relationships), with the inputs
and output being random variables, and the reasoning steps as a mediator
variable. Extending this formulation by \citet{paul-etal-2024-making}---who
treated the entire reasoning trace as a single atomic mediator---we posit that
our constructed \textbf{CCGraph} $G$ can be treated as a ``fine-grained'' causal
graph, which models the relationship between the inputs ($\hat{q} \in \hat{Q}$)
and the output ($\hat{a}$), mediated by terms in the reasoning trace
($\hat{r} \in \hat{R}$) in accordance with the DAG $G$.

We can then try to assess the \emph{direct effect} (DE), i.e., how an
intervention on $\hat{q}$ affects $\hat{a}$, without passing through the
mediators $\hat{r} \in \hat{R}$, and the \emph{indirect effect} (IE), i.e., how
an intervention affects $\hat{a}$, indirectly through $\hat{R}$. Critically, if
for a given LLM we find that there is no indirect effect, i.e., \emph{the final
answer has no causal contributions from terms in the provided reasoning trace}
we can conclude that the \emph{CoT} traces it produces are mere decoration, and
the mechanism through which \emph{CoT} improves performance \textbf{is not
reasoning}. This serves as the central framework for analysis in the rest of
this paper.

\noindent\textbf{Attention Suppression:} To investigate reasoning node mediation
we employ the ``\emph{attention suppression}'' intervention
\citep{bogdan2025thoughtanchors}. Given a sequence of tokens $X = \{x_1, \ldots
, x_T\}$ and $X_{\text{suppressed}} \subseteq X$, the modified attention is
defined as:
\begin{align}
    A_i^{(\phi)}  &= W^O \cdot \textrm{concat}(A_i^{1(\phi)}, \ldots A_i^{\textrm{\# head}(\phi)})\\
    A_i^{j(\phi)} &= \sum_{\substack{k=1\\x_k \notin X_{\text{supp.}} }}^{T} \textrm{sim}(Q_i^{j(\phi)}, K_k^{j(\phi)}) \cdot V_k^{j(\phi)}
\end{align}
where $j \in \{1, \ldots, \textrm{\# heads}\}$, and $\phi \in \{1, \ldots,
\textrm{\# layers}\}$. $Q_i^{j(\phi)}$, $K_i^{j(\phi)}$, and
$V_i^{j(\phi)}$ are the $i^{\text{th}}$ token query, key and value projections
for the $j^{\text{th}}$ head and $\phi^{\text{th}}$ layer, respectively. This
zeros out the effect of the suppressed tokens (across all layers and heads) and
ensures \emph{that all information flow from the suppressed tokens is cut off},
i.e., these tokens have no influence on any other tokens. We apply this
intervention on tokens of the reasoning trace to model the counterfactual
scenario where these tokens were absent from the reasoning trace.
\citet{bogdan2025thoughtanchors} further asserted that this approach
\emph{``...does not problematically induce out-of-distribution behavior.''}

\section{Experiments}
\label{sec:results}

\subsection{Is CCGraph a Mediator?}
\label{subsec:mediate}

\begin{table*}
    \begin{center}
        \footnotesize{
        \begin{tabular}[c]{l|l c r|l c r|l c r}
            \toprule
            \multirow{3}{*}{Model}  & \multicolumn{3}{c|}{\textbf{GSM8K}}                     & \multicolumn{3}{c|}{\textbf{MATH500}}                   & \multicolumn{3}{c}{\textbf{AIME}}                      \\
            \cmidrule{2-10}
                                    & Orig.              & \multicolumn{2}{c|}{\textbf{AS}}   & Orig.              & \multicolumn{2}{c|}{\textbf{AS}}   & Orig.              & \multicolumn{2}{c}{\textbf{AS}}   \\ 
                                    & $H(P_A)$           & $H(P_A^M)$ & $D_{KS}$              & $H(P_A)$           & $H(P_A^M)$ & $D_{KS}$              & $H(P_A)$           & $H(P_A^M)$ & $D_{KS}$             \\
            \midrule
            DeepSeek R1 1.5B        & 0.02               & 3.58       & 1.00                  & 0.46               & 1.37       & 0.68                  & 0.04               & 1.31       & 0.95                 \\
            Qwen 3 1.7B             & $1\mathrm{e}{-3}$  & 0.85       & 0.99                  & 0.08               & 0.38       & 0.70                  & $5\mathrm{e}{-4}$  & 0.39       & 0.95                 \\
            Gemma 3 1B              & $2\mathrm{e}{-3}$  & 0.88       & 0.98                  & 0.09               & 0.49       & 0.74                  & 0.01               & 0.49       & 0.92                 \\
            \midrule
            Llama 3.1 8B            & $3\mathrm{e}{-3}$  & 3.23       & 0.99                  & 0.36               & 1.28       & 0.49                  & 0.05               & 1.56       & 0.61                 \\
            Qwen 2.5 7B             & $1\mathrm{e}{-3}$  & 1.24       & 0.99                  & 0.16               & 0.58       & 0.55                  & 0.01               & 0.45       & 0.82                 \\
            Qwen 2.5 7B Math        & 0.08               & 3.30       & 0.99                  & 0.58               & 1.48       & 0.63                  & 0.13               & 1.05       & 0.89                 \\
            DeepSeek R1 8B          & 0.04               & 3.69       & 0.99                  & 0.28               & 1.79       & 0.68                  & 0.04               & 2.53       & 0.92                 \\
            DeepSeek R1 8B (0528)   & 0.01               & 2.39       & 0.99                  & 0.29               & 1.01       & 0.67                  & 0.02               & 1.27       & 0.98                 \\
            Qwen 3 8B               & $8\mathrm{e}{-4}$  & 0.99       & 0.99                  & 0.15               & 0.41       & 0.61                  & $3\mathrm{e}{-4}$  & 0.34       & 0.92                 \\
            Gemma 3 12B             & $7\mathrm{e}{-4}$  & 1.02       & 0.99                  & 0.08               & 0.48       & 0.69                  & 0.01               & 0.46       & 0.87                 \\
            \midrule
            DeepSeek R1 32B         & $6\mathrm{e}{-3}$  & 2.73       & 0.99                  & 0.36               & 1.36       & 0.73                  & 0.02               & 1.14       & 0.97                 \\
            Qwen 3 32B              & $2\mathrm{e}{-3}$  & 1.02       & 0.99                  & 0.15               & 0.59       & 0.67                  & 0.01               & 0.43       & 0.92                 \\
            Gemma 3 27B             & $4\mathrm{e}{-4}$  & 0.27       & 0.99                  & 0.08               & 0.27       & 0.63                  & $2\mathrm{e}{-3}$  & 0.22       & 0.82                 \\
            Llama 3.3 70B           & $2\mathrm{e}{-4}$  & 0.91       & 0.97                  & 0.15               & 0.40       & 0.42                  & 0.09               & 0.35       & 0.31                 \\
            DeepSeek R1 70B         & 0.02               & 3.92       & 0.99                  & 0.40               & 1.53       & 0.65                  & 0.04               & 1.77       & 0.89                 \\
            \bottomrule
        \end{tabular}}
    \end{center}
    \caption{\textbf{Are reasoning nodes mediators?} $H(P_A)$ refers to the
    entropy of the (first token of the) answer, averaged over the population of
    problems. $D_{KS}$ refers to the Kolmogorov distance between the original
    (Orig.) and attention-suppressed (AS) distributions of entropy, measured
    with a 2-sample KS test. We find that attention suppression for tokens
    corresponding to every reasoning node in $G$ increases answer uncertainty
    significantly ($p < 10^{-12}$).
    }
    \label{tab:graph-intervene}
\end{table*}

We first test whether the reasoning nodes in the \textbf{CCGraph} serve as
mediators for the final answer by performing \emph{attention suppression} on all
reasoning nodes in a \textbf{CCGraph} for a problem. Our results, summarized in
Table \ref{tab:graph-intervene}, show that attention suppression over all tokens
corresponding to every reasoning node in a \textbf{CCGraph} significantly
increases uncertainty in the answer ($p < 10^{-12}$). The answer entropy,
defined as:
\begin{align*}
    &H(P_t) = -\!\sum_{v \in V}\!P(x_t\!=\!v|x_{<t}) \log P(x_t\!=\!v| x_{<t})\\
    &P_A = P(A_0|\, x_{<T}) \tag*{$x_{<T}\rightarrow$ tokens before $A_0$}\\
    &P_A^M = P(A_0|\,x_T, \ldots, x_{\gamma+1}, x_{\gamma-\delta-1}, \ldots)
\end{align*}
where $x_{\gamma-\delta}, \ldots, x_{\gamma}$ are tokens corresponding to an
attention-suppressed reasoning node ($\hat{r} \in G$), and $A_0$ represents the
first token of the answer. The large Kolmogorov distance $D_{KS}$, measured with
a 2-sample Kolmogorov-Smirnov (KS) test, indicates a significant shift in the
distribution of $H(P_A)$ between the original and attention-suppressed cases,
leading us to conclude that answers ($\hat{a}$) are mediated by reasoning nodes,
a necessary condition for reasoning.

\subsection{Validating R Paths via Counterfactuals}
\label{subsec:rpath}

\begin{figure*}[!t]
    \begin{center}
        \includegraphics[width=\textwidth]{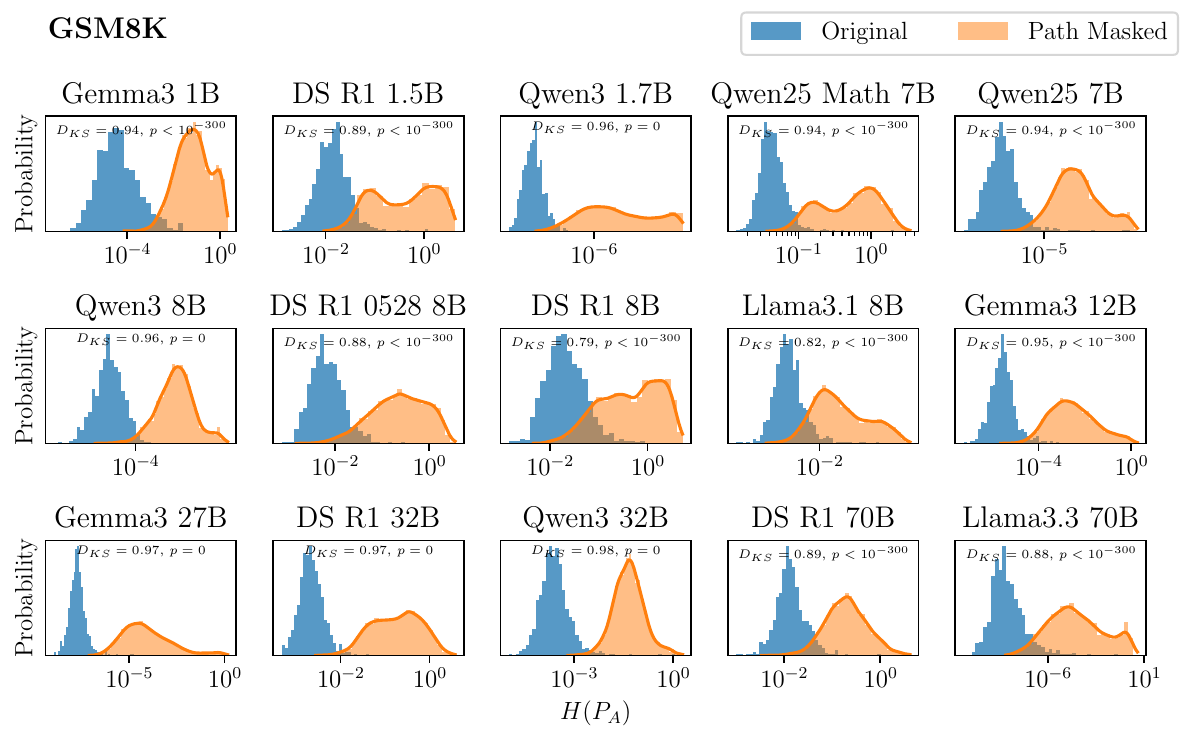}
    \end{center}
    \caption{\textbf{Do reasoning path interventions affect the answer?}\\
        We find that when attentions corresponding to tokens in an \textbf{R
        path} are suppressed, the entropy of the distribution of the answer
        ($H(P_A)$) increases significantly, i.e., uncertainty over the answer is
        significantly increased. The figure also reports results of the 2-sample
        KS test, showing high values of Kolmogorov distance ($D_{KS}$) and high
        statistical significance ($p < 10^{-300}$).
    }
    \label{fig:path-mask}
\end{figure*}

With our \textbf{R paths} extracted from the \textbf{CCGraphs} constructed for
each sample, we perform attention suppression on tokens corresponding to reasoning
nodes in an \textbf{R path}.\footnote{$\hat{q}$ nodes are not suppressed.} As an
illustrative example, the nodes marked in \emph{blue} in Figure
\ref{fig:example_annotation} \emph{(right)} are suppressed, and the effect of
the intervention on the distribution for the answer node (\emph{green}) is
examined.

Figure \ref{fig:path-mask} summarizes the findings for our experiment with the
GSM8K split of \textbf{KisMATH} (results with other splits are presented in
Appendix \ref{sec:appendix-additional}, Figure \ref{fig:path-mask-2}). We find
that the \emph{attention suppression intervention significantly increases the
uncertainty in the answer} (see Figure \ref{fig:path-mask}). A 2-sample KS test
with samples of answer entropy ($H(P_A)$) from the original and intervened
distribution shows high values of Kolmogorov distance ($D_{KS}$), and extremely
low $p$-values. This leads us to \emph{reject the null hypothesis} and conclude
that \textbf{R path} suppression has a significant effect on reasoning outcome.

\subsection{Realization of Causal Structure}

To assess whether language models (LMs) internally realize a structure similar
to the \textbf{CCGraph}, we check whether they emphasize reasoning paths from
the \textbf{CCGraph}. We measured the probability assigned to an \textbf{R path}
$\mathcal{R} = [\hat{q}_{\alpha} \rightarrow \hat{r}_{(i_1)} \rightarrow \hat{r}_{(i_2)} \rightarrow \ldots \hat{r}_{(i_{\mu})} \rightarrow \hat{a}]$
which is defined as:
\begin{align}
    P(\mathcal{R}) &= \prod_{\delta = 1}^{\mu} P \Big( \hat{r}_{(i_{\delta})}   \Big| x_{<T_{\delta}} \Big)\\
    P(\hat{r}_{(i_{\delta})}| x_{<T_{\delta}}) &=
        \prod_{\lambda = 1}^{n} P\Big(t_{\lambda}^{\delta}\Big|t_{\lambda - 1}^{\delta}, \ldots t_1^{\delta}, x_{<T_{\delta}}\Big)
    \label{eq:path-prob}
\end{align}
where $t_{1}^{\delta}, \ldots t_{n}^{\delta}$ are tokens of $\hat{r}_{(i_{\delta})}$,
and $x_{<T_{\delta}}$ represents every token appearing before the reasoning node
$\hat{r}_{(i_{\delta})}$. This is \emph{not the exact path transition
probability}, as that would require marginalizing over the distribution of all
intermediate tokens, which is intractable. To sidestep this problem, we compare
probabilities of \textbf{R paths} with respect to $M$ random paths
$\tilde{\mathcal{R}}_{\kappa}$, which are not a part of the \textbf{CCGraph} and
report the distribution of the rank of $P(\mathcal{R})$ (how many random paths
is an \textbf{R path} $\mathcal{R}$ better than):
\begin{equation}
    \text{rank}_M(\mathcal{R}) = \frac{1}{M}\sum_{\kappa = 1}^{M}\mathbb{I}\Big[ P(\mathcal{R}) > P(\tilde{\mathcal{R}}_{\kappa})\Big] 
    \label{eq:rank}
\end{equation}
In our analysis, a random path $\tilde{\mathcal{R}}_{\kappa}$ is constructed by
randomly choosing an \textbf{identical number of tokens} (same length) as the
\textbf{R path} from the \textbf{same reasoning trace} avoiding \textbf{CCGraph}
nodes (see Figure \ref{fig:example_annotation} for an example). We then compute
$P(\mathcal{R})$ and $P(\tilde{\mathcal{R}}_{\kappa})$ for all $\kappa$ (see Eq.
\ref{eq:path-prob}) and report the distribution of ranks. We only examine the
relative rank of the path probability $P(\mathcal{R})$ with respect to the path
probability of several other equal-length random paths
($P(\tilde{\mathcal{R}}_{\kappa})$) to control for variations in path
probabilities arising out of differences in length, token frequency, etc.

\begin{figure*}[t!]
    \begin{center}
        \includegraphics[width=\textwidth]{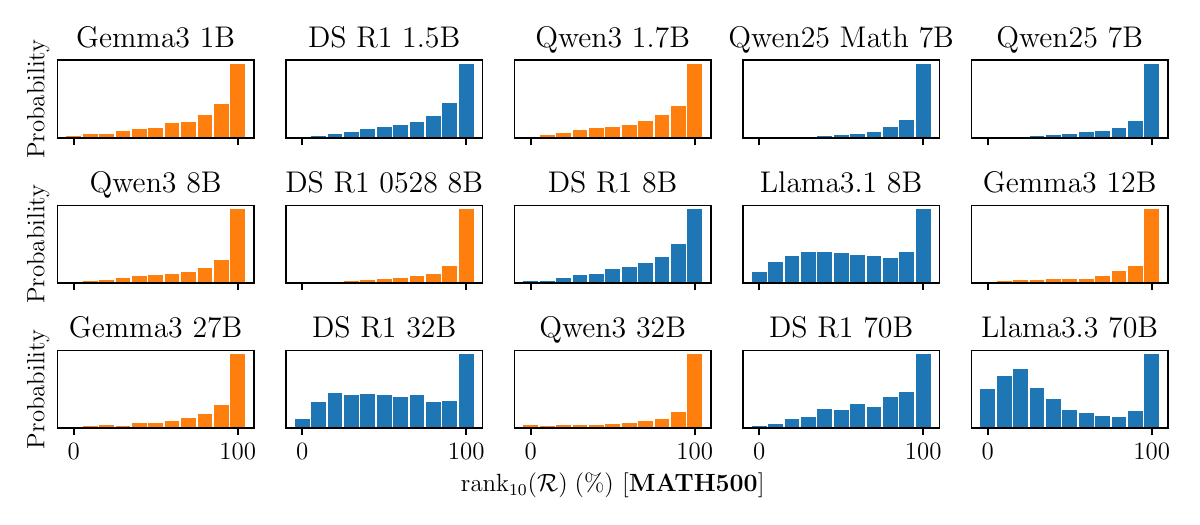}
    \end{center}
    \caption{\textbf{Are LLMs aware of implicit structures in reasoning?}\\
    We compare the probability associated with \textbf{reasoning paths} (see Eq.
    \ref{eq:path-prob}) with the probability of a random path through the
    reasoning response (e.g. Figure \ref{fig:example_annotation} \emph{(right)}).
    The graphs show the rank of a reasoning path compared to random paths (see
    Eq. \ref{eq:rank}) for various models. A striking peak is observed at the
    100 \%-ile region, indicating that a large fraction of reasoning paths
    entirely consist of higher probability transitions.
    }
    \label{fig:path-prob}
\end{figure*}

Our results with the MATH500 split of \textbf{KisMATH} are summarized in Figure
\ref{fig:path-prob} (AIME, GSM8K splits in Appendix \ref{sec:appendix-additional},
Figure \ref{fig:path-prob-2}). Our observations are as follows:

\noindent $\blacklozenge$ \textbf{Pronounced spike at the 100th percentile:} All
tested LLMs across the 3 different splits show a spike at 100 percentile
(barring Llama 3.3 70B Instruct on AIME), which indicates that $P(\mathcal{R})$
(Eq. \ref{eq:path-prob}) is high for a considerable fraction of
\textbf{R paths}. This property requires that these LLMs produce
higher-probability transitions for almost all tokens along \textbf{R paths} when
compared to random paths $\tilde{\mathcal{R}} \notin G$. This hints at the fact
that a structure similar to the proposed \textbf{CCGraph} is implicitly realized
in \emph{CoT} traces.

\noindent $\blacklozenge$ \textbf{Two behavior modes:} From the rank
distributions in Figures \ref{fig:path-prob} and \ref{fig:path-prob-2}, we see
two patterns of behavior, a ``\textbf{bell}''-shaped and an
``\textbf{exponential}''-shaped distribution (or a combination of both). The
\textbf{exponential}-shaped behavior is most prevalent across all the models
tested and is to be expected if \emph{reasoning paths always correspond to
high-probability transitions}; i.e., $P(\mathcal{R})$ is always high for
\textbf{R paths}. The ``\textbf{bell}''-shaped distribution indicates that a
small fraction of \textbf{R paths} contain low-probability (high-entropy)
transitions, which is consistent with recent findings by
\citet{wang20258020rulehighentropy}. We discuss this further in Section
\ref{subsec:wide}.

\section{Discussions and Further Analysis}
\label{sec:analysis}

\subsection{Is Math All You Need?}
\label{subsec:mayn}

Based on the results in Section \ref{sec:results}, which show that mathematical
entities in a \emph{CoT} trace are mediators for the final outcome, and that LLM
distributions are aligned to the relationships between these entities, a natural
followup question is---\emph{does the context surrounding mathematical entities
matter?}

To investigate this we tested LLMs with two interventions: $M(G)$ where
attention-suppression was performed on tokens corresponding to reasoning nodes
in a \emph{CoT} trace (similar to experiment in Section \ref{subsec:mediate}),
and $M(G^C)$ where attention-suppression was performed on all tokens which are
not part of any reasoning node in a \emph{CoT} trace. The intervention $M(G)$
only allows the final answer to glean information from natural language text
surrounding mathematical expressions, whereas $M(G^C)$ only allows information
from mathematical expressions. %(e.g., for Figure \ref{fig:example_annotation} $M(G^C)$ results in ``\texttt{...4, 25\%, 4, $0.25\times4=1$, $4+1=5$, $4+5=9$...}'').
We measured \emph{Answer Change} (\%), i.e., how often does the (first-token of
the) answer change in response to these interventions, and our results are
summarized in Table \ref{tab:graph-intervene-acc} in Appendix \ref{sec:appendix-additional}.

Our results indicate that the intervention $M(G)$ (math-suppressed) changes the
final answer much more reliably ($70.9\%$ vs. $10.3\%$ on average) compared to
$M(G^C)$ (math-only) for the GSM8K dataset, suggesting that the mathematical
expressions play a more significant role. However, for more involved problems
from the MATH500 and AIME datasets, where the natural language ``glue'' might
contain non-trivial deductive reasoning, such a clear pattern is not observed.
$M(G)$ has a slightly stronger effect on the final answer, but a statistically
significant difference ($\chi^2$-test, $\alpha=0.01$) between the two
interventions is not observed in most cases.

From a qualitative overview, we observe that the $G^C$ in the \emph{CoT} trace
in the more complex problems contains a number of what linguists have called
\emph{discourse connectives} \citep{rst,sdrt,pdtb, nicholas2}, as well as
strings that help specify the role of the elements in $G$. Discourse connectives
are important for determining the discourse structure of a text, which in turn
tells us how the entities in $G$ are semantically related. Ideally, both are
needed to reflect reasoning. Linguists have used entity chains, which resemble
our \textbf{R paths} as an approximation for discourse structure \citep{oberlander}.
The chains themselves are not sufficient to determine discourse structure but in
fact presuppose it and reflect it, as one chain can be extracted from many texts
with heterogeneous meaning and structure. Nevertheless, when the texts have a
simple and largely fixed discourse structure as in GSM8K, then the models will
have learned the discourse structure and the chain in $G$ is sufficient for
reasoning.  When the \emph{CoT} trace is more complex and has a more elaborate
discourse structure as in AIME or MATH500, then the model needs the discourse
structure and the semantic relations between the entities in $G$ to reason. This
could account for the results presented in Table \ref{tab:graph-intervene-acc}.

\subsection{LMs with a Wide Distribution}
\label{subsec:wide}

\begin{figure}[!t]
    \begin{center}
        \includegraphics[width=0.5\textwidth]{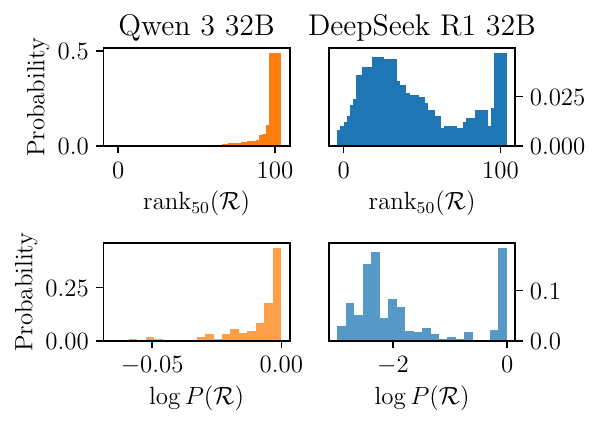}
    \end{center}
    \caption{
        \textbf{Analyzing the ``bell''-shape.}
        We compare a higher-resolution \textbf{R path} rank-distribution
        ($\text{rank}_{50}(\mathcal{R})$) for two models exhibiting behavior on
        the two ends of the spectrum of rank distributions (see Figure
        \ref{fig:path-prob-2}). The model demonstrating ``bell''-shape (DeepSeek
        R1 32B) has lower $P(\mathcal{R})$ scores for some \textbf{R paths} and the
        scores have higher variance. Results are reported with 100 samples from the
        AIME split.
    }
    \label{fig:high-res-aime}
\end{figure}

In Section \ref{subsec:rpath}, we noted the two behavior regimes demonstrated by
LMs at the \textbf{R path}-awareness task, which we investigate further in this
section. Figure~\ref{fig:high-res-aime} compares two models demonstrating
behavior on the extreme ends of the spectrum between the ``bell''
(DeepSeek R1 32B) and ``exponential'' (Qwen3 32B) shape. We plot a
higher-resolution version of the rank-distribution ($\text{rank}_{50}(\mathcal{R})$)
and $\log P(\mathcal{R})$ (see Eq. \ref{eq:path-prob}) corresponding to
\textbf{R paths} from the AIME split.

Our analysis shows that the ``bell''-shaped curve results from models generating
a small fraction of low-probability transitions along some \textbf{R paths}. The
sum of transition $logprobs$ ($\log P(\mathcal{R})$) along \textbf{R paths}, in
addition to being on average lower ($-1.7603$ for DeepSeek R1 32B vs.~$-0.0098$
for Qwen3 32B), demonstrates higher variance ($0.9217$ for DeepSeek R1 32B
vs.~$0.0002$ for Qwen3 32B).

\begin{figure}[!t]
    \begin{center}
        \includegraphics[width=0.5\textwidth]{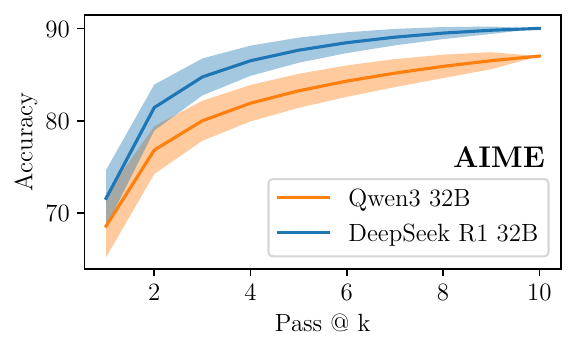}
    \end{center}
    \caption{\textbf{High entropy along R paths drive exploration.}
    The graph presents performance ($pass@k$-accuracy) of the two models
    (DeepSeek R1 32B, Qwen3 32B) with varying number of sample rollouts ($k$).
    The DeepSeek R1 32B model has higher uncertainty (lower probability
    transitions for reasoning paths, and higher variance), which might enable a
    more thorough exploration of diverse reasoning paths.
    }
    \label{fig:perf-var}
\end{figure}

This enhanced uncertainty during a constrained traversal along an \textbf{R path}
can arise for two reasons: (i) \emph{intrinsic uncertainty}, i.e., uncertainty
inherent in the reasoning process; and (ii) \emph{extrinsic uncertainty}, i.e.,
uncertainty arising from external factors not inherent to reasoning.
\emph{Intrinsic uncertainty} refers to the fact that the reasoning process has
some underspecification, such as names for bound variables, order in which
sub-problems/lemmas are solved, and even potential disparate approaches leading
to the correct solution. In contrast, \emph{extrinsic uncertainty} may arise
from poor modeling of the reasoning process, hallucination propensity, etc. If
LMs model \emph{intrinsic uncertainty} we should expect to see a more thorough
exploration of alternative valid solution paths, i.e., sampling a higher number
of \emph{CoT} rollouts should improve performance.

To study this, we compare $pass@k$-performance of a ``bell''-shape model
(DeepSeek R1 32B) and an ``exponential''-shape model (Qwen 3 32B) with
increasing numbers of sampled rollouts (see Figure~\ref{fig:perf-var}). $pass@k$
considers an answer correct if any of the $k$ generated samples is correct, and
thus indicates whether a greater proportion of explored solutions are correct.
The performance of the two models is close at low sample counts ($68.6\% \pm 3.4$
for Qwen3 32B vs.~$71.6\% \pm 3.0$ for DeepSeek R1 32B with $k=1$) and it widens
with increasing sample count ($87\%$ for Qwen3 32B vs.~$90\%$ for DeepSeek 32B
with $k=10$). The ``bell''-shaped model (DeepSeek R1 32B) indeed shows improved
performance indicating that it explores a greater variety of reasoning paths.
This finding concurs with related findings by \citet{wang20258020rulehighentropy},
who found that a small fraction of tokens exhibit high entropy and act as
``forks'' enabling more robust exploration of diverse reasoning paths.

Additionally, note that DeepSeek R1 32B was created by distilling \emph{CoT}
rollouts from DeepSeek R1 671B \citep{deepseek-r1}, whereas Qwen3 32B underwent
reinforcement learning with verifiable rewards (RLVR) post-training
\citep{qwen3}. Our observations with regard to RLVR-induced ``over-confidence'',
are in line with recent findings by \citet{yue2025doesreinforce}, which showed
that the base model at high sample counts outperforms its RLVR-trained
counterpart, and suggest that RLVR post-training worsens exploration.

\section{Conclusions}
\label{sec:conclusion}

In this work we examine the role that LLM generated \emph{chain-of-thought}
traces play in mathematical reasoning through a causal lens. To this end, we
proposed a procedure that recovers a \emph{directed-acyclic graph} from a
\emph{CoT} trace, whose nodes are mathematical expressions mentioned in the
trace and whose edges encode fine-grained causal links between them. Using this
procedure we create \textbf{KisMATH}---a large-scale collection of problems
paired with their LLM solutions and ``\emph{Causal CoT graphs}''
(\textbf{CCGraphs}). \textbf{KisMATH} enables us to perform interventions on
LLMs in a controlled, graph-aware manner for multi-step mathematical reasoning
problems, which can be more informative than stochastic interventions.

Mediation by the reasoning nodes in the \textbf{CCGraphs} implicit in the
\emph{CoT} trace is a \emph{necessary} condition for reasoning, i.e., if
reasoning nodes in the \textbf{CCGraphs} have no \emph{indirect effect} and
thus, do not act as mediators between the question and the generated answer, we
can conclude that LLMs do not reason. However, our experiments with 15 popular
state-of-the-art open-weight LLMs ranging from 1B--70B parameters, consistently
find strong mediation by reasoning nodes, and removing them significantly
increases answer entropy. Further, comparing the probability of traversing
through the reasoning trace using \textbf{CCGraph}-aligned ``reasoning paths''
with that of random paths, we observe that reasoning paths consistently receive
a higher probability mass. Additional analysis with the aid of \textbf{CCGraphs}
reveals that model behavior broadly falls into two regimes: an ``exponential''
regime where almost every reasoning path consists of high-probability
transitions, and a ``bell-shaped'' regime where a minority of low-probability
``fork'' tokens enable broader exploration.

Our findings suggest that intermediate reasoning tokens serve a crucial role in
arriving at the answer to mathematical-reasoning problems, and LLMs internally
favour the same paths that our graph extraction procedure identifies, thus
outlining that structures implicit to reasoning are embedded in reasoning traces.
We hope that \textbf{KisMATH} facilitates further research into uncovering latent
structures present in LLM reasoning traces.

% ----------------------- BACKMATTER -------------------------------------------
\section*{Limitations}
\label{sec:limit}

\textbf{Manual Intervention:} Algorithm \ref{alg:graph-construct}
starts by modeling relationships between mathematical expressions and involves
a pruning step. Since we are interested in investigating the indirect effect of
intermediate reasoning tokens, the pruning step removes nodes from the
\textbf{CCGraph} which do not lie on a path from a question node to the answer
node. In some instances ($\sim10\%$), if a few critical links between the
question and answer are not captured, the \textbf{CCGraph} degenerates into a
trivial graph containing only the answer node. This was addressed by manual
intervention as described in Section \ref{subsec:ccgc}.

\textbf{Scope:} Our approach to extracting \textbf{CCGraphs} is not
directly applicable to some reasoning problems, such as geometry, abstract
mathematics, commonsense reasoning, etc. For example, consider the statement -
\emph{"Let the midpoint of line $\overline{AB}$ be $O$..."} or \emph{"Let $H$ be
a normal subgroup of group $G$..."}. In such settings it is difficult to decipher
the relationship between the entities from algebraic parsing alone, and invoking
our analysis techniques would require additional inputs. Future work geared
toward developing techniques to capture semantic similarity between such
entities by incorporating natural language context would be compelling.

\section*{Ethics Statement}

In keeping with ACL ethical guidelines, all scientific artifacts generated for
this study---including code, prompts, data, and raw model outputs---are made
freely available as open source under the MIT license. Only public datasets
available on the \href{https://huggingface.co/}{Huggingface} platform were used
in the study, and the usage of these datasets is in accordance with their
intended usage. Beyond minimal use in writing (e.g., grammar suggestions,
finding synonyms), AI assistants were not used in ideation, coding, or writing
involved in this work. We foresee no serious ethical implications for society at
large from this study.

\section*{Acknowledgments}

The authors are grateful to the reviewers and the action editor for their
insightful feedback, which was instrumental in shaping this manuscript. The
authors would also like to extend their gratitude to OpenAI for providing a
computational grant under their researcher access program (\# \texttt{0000010960}).
This research was funded in part by the Indo-French Centre for the Promotion of
Advanced Research (IFCPAR/CEFIPRA) through project number \texttt{CSRP 6702-2}.
This work was also supported by BPI France through the project OpenLLM France
(\texttt{DOS0250771} and \texttt{DOS0250773}).

% ----------------------- REFERENCES -------------------------------------------
%\clearpage
\bibliographystyle{requirements/acl_natbib.bst}
\bibliography{backmatter/references.bib, backmatter/model_references.bib}
%\clearpage
% ----------------------- APPENDICES -------------------------------------------
\appendix
\section{Appendix: Additional Data}
\label{sec:appendix-additional}

\begin{figure*}[!t]
    \begin{center}
        \begin{subfigure}[t]{\textwidth}
            \includegraphics[width=\textwidth]{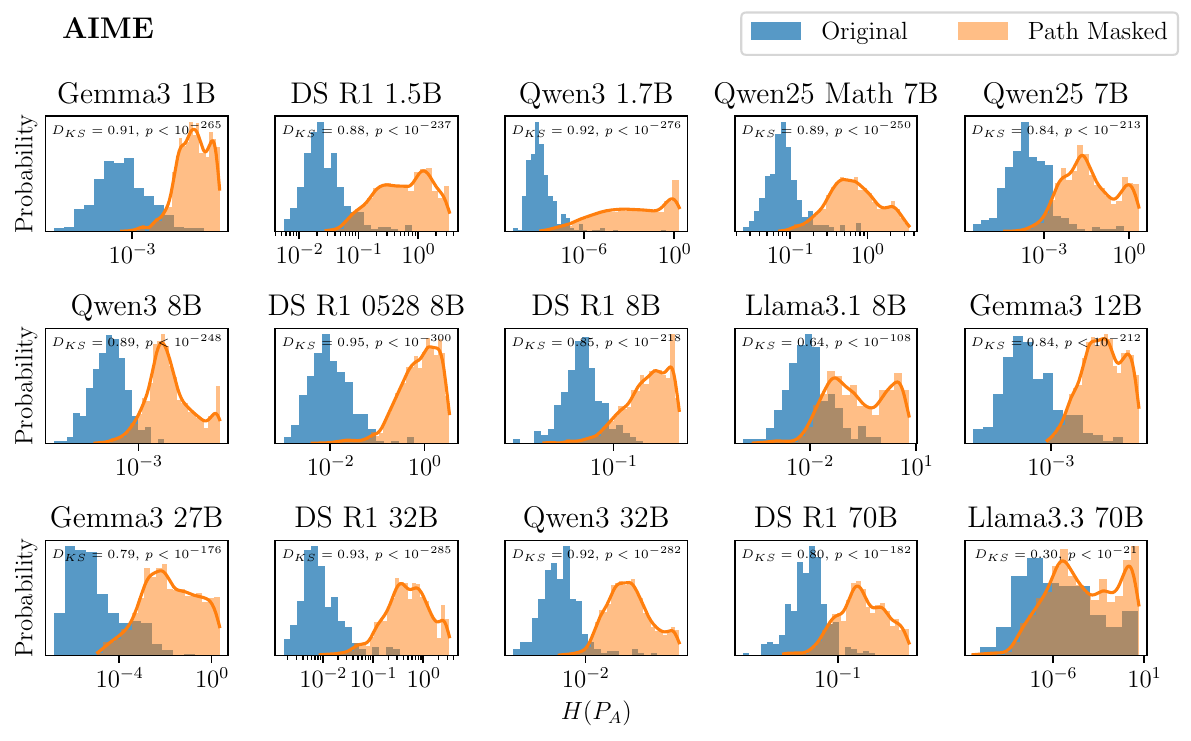}
        \end{subfigure}
        \begin{subfigure}[b]{\textwidth}
            \includegraphics[width=\textwidth]{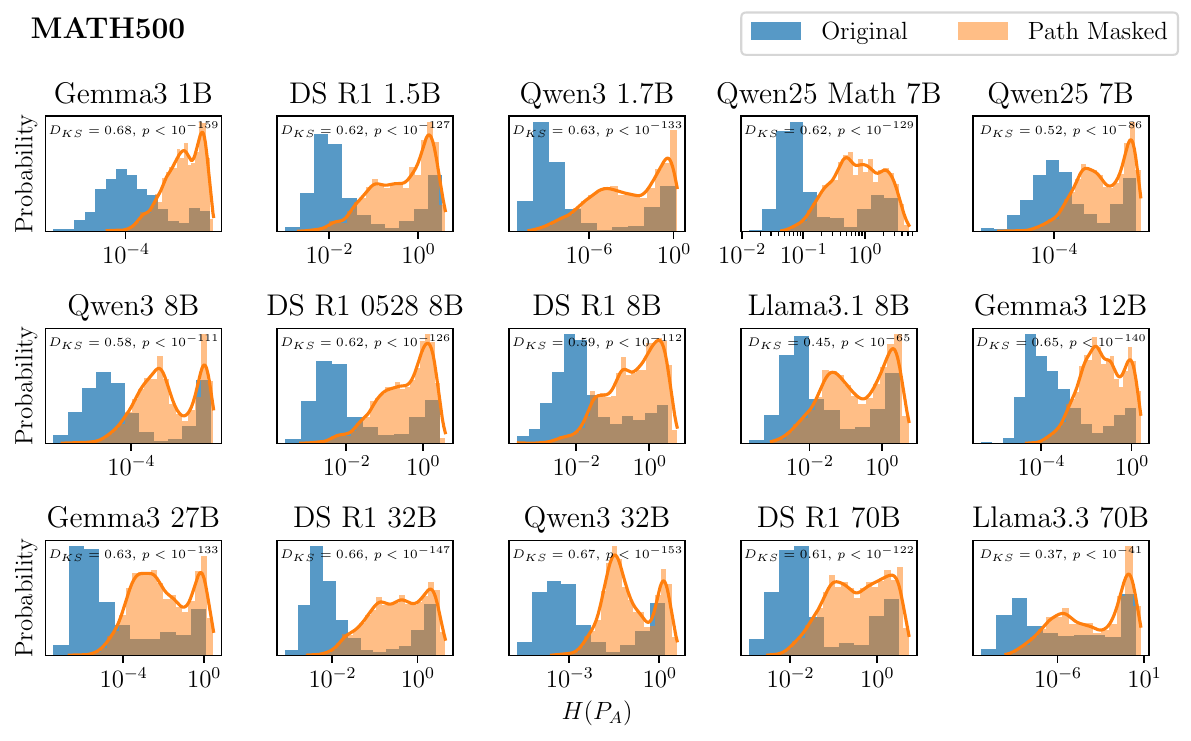}
        \end{subfigure}
    \end{center}
    \caption{Results with additional splits for experiment in Section
    \ref{sec:method}, Figure \ref{fig:path-mask}.}
    \label{fig:path-mask-2}
\end{figure*}

\begin{figure*}[!t]
    \begin{center}
        \begin{subfigure}[t]{\textwidth}
            \includegraphics[width=\textwidth]{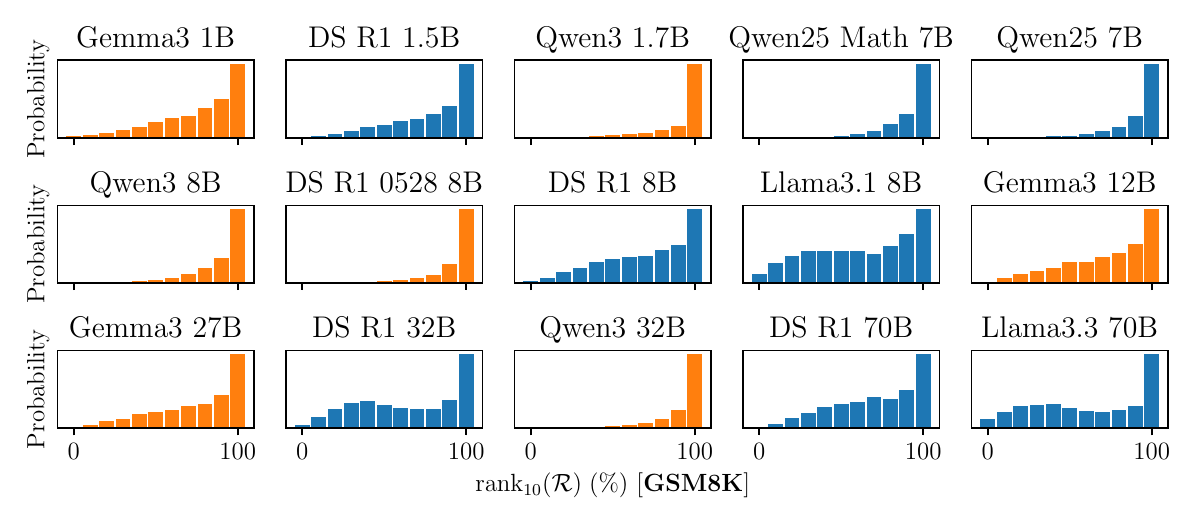}
        \end{subfigure}
        \begin{subfigure}[b]{\textwidth}
            \includegraphics[width=\textwidth]{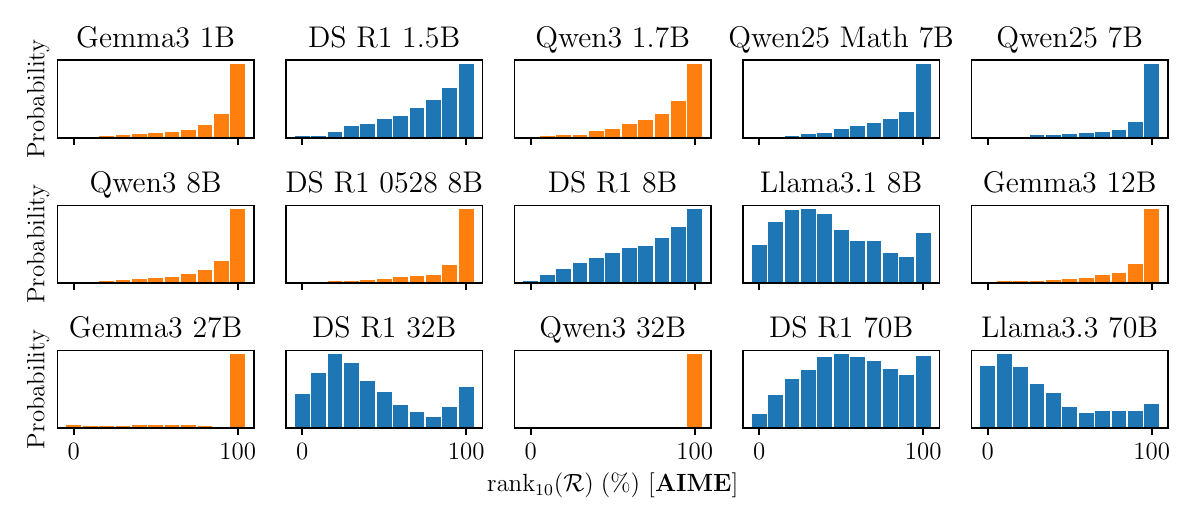}
        \end{subfigure}
    \end{center}
    \caption{Results with additional splits for experiment in Section
    \ref{sec:results}, Figure \ref{fig:path-prob}.}
    \label{fig:path-prob-2}
\end{figure*}

Please see Figure \ref{fig:path-mask-2} and \ref{fig:path-prob-2} for
experimental results of Section \ref{sec:results} with additional dataset
splits. Table \ref{tab:graph-intervene-acc} contains results for experiment in
Section \ref{subsec:mayn}

%\subsection{Approximate Retrieval}
%\label{sec:aretrieval}
%
%We consider if \emph{CoT} traces merely enable LMs to perform approximate
%retrieval from its latent knowledge. In this scenario it is expected that
%attention suppression on ``retrieval query'' tokens would lead to a change in
%answers. To investigate this we performed two
%interventions---$M(G)$, where all reasoning tokens corresponding to \textbf{CCG}
%$G$ is suppressed (similar to experiment in Table \ref{tab:graph-intervene}),
%and $M(G^C)$, where all reasoning tokens that are not constituents of any
%$\hat{r} \in G$ are suppressed. Suppressing mathematical expressions should have
%a significantly stronger effect on the answer as opposed to other tokens, and
%we indeed observe that same for the GSM8K split. However, this pattern is absent
%from the AIME and MATH500 splits, which feature much more complex problems,
%often requiring significant verbal reasoning alongside algebraic manipulation.
%The observations, summarized in Table \ref{tab:graph-intervene-acc}, better
%align with the hypothesis that LMs reason, and suppression of the trivial
%connective glue between mathematical expressions in GSM8K is not sufficient to
%cause answer change, but suppression of natural language mathematical reasoning
%or mathematical expressions have similar effects for MATH500 and AIME.

\begin{table*}
    \begin{center}
        \footnotesize{
        \begin{tabular}[c]{l|l r|l r|l r}
            \toprule
            \multirow{3}{*}{\textbf{Model}}  & \multicolumn{6}{c }{Answer Changed $(\%)$}        \\
            \cmidrule{2-7}
                                    & \multicolumn{2}{c|}{\textbf{GSM8K}}  & \multicolumn{2}{c|}{\textbf{MATH500}}     & \multicolumn{2}{c}{\textbf{AIME}}     \\
                                    & $M(G)$          & $M(G^C)$           & $M(G)$      & $M(G^C)$                    & $M(G)$      & $M(G^C)$                \\
            \midrule
            DeepSeek R1 1.5B        & \textbf{96.2}   & \textbf{22.6}      & 43          & 42                          & 38          & 43                      \\
            Qwen 3 1.7B             & \textbf{72.7}   & \textbf{9.8}       & 42          & 34                          & 40          & 33                      \\
            Gemma3 1B               & \textbf{85.5}   & \textbf{69.8}      & 68          & 63                          & 78          & 79                      \\
            \midrule
            Llama 3.1 8B            & \textbf{81.2}   & \textbf{1.9}       & 43          & 39                          & 33          & 39                      \\
            Qwen 2.5 7B             & \textbf{61.2}   & \textbf{4.7}       & 37          & 25                          & 24          & 19                      \\
            Qwen 2.5 7B Math        & \textbf{64.1}   & \textbf{2.1}       & \textbf{27} & \textbf{13}                 & 18          & 14                      \\
            DeepSeek R1 8B          & \textbf{94.0}   & \textbf{9.1}       & 49          & 37                          & \textbf{56} & \textbf{37}             \\
            DeepSeek R1 8B (0528)   & \textbf{69.9}   & \textbf{3.2}       & 34          & 24                          & 31          & 24                      \\
            Qwen 3 8B               & \textbf{53.7}   & \textbf{2.4}       & \textbf{31} & \textbf{19}                 & 23          & 20                      \\
            Gemma 3 12B             & \textbf{58.1}   & \textbf{14.5}      & 33          & 30                          & 32          & 38                      \\
            \midrule
            DeepSeek R1 32B         & \textbf{91.5}   & \textbf{1.3}       & 28          & 18                          & 22          & 16                      \\
            Qwen 3 32B              & \textbf{39.0}   & \textbf{0.7}       & 25          & 20                          & 13          & 12                      \\
            Gemma 3 27B             & \textbf{49.6}   & \textbf{10.6}      & 27          & 23                          & \textbf{22} & \textbf{38}             \\
            Llama 3.3 70B           & \textbf{62.3}   & \textbf{1.6}       & 40          & 33                          & 21          & 27                      \\
            DeepSeek R1 70B         & \textbf{84.5}   & \textbf{0.9}       & \textbf{39} & \textbf{19}                 & \textbf{30} & \textbf{12}             \\
            \bottomrule
        \end{tabular}}
    \end{center}
    \caption{Results pertaining to discussion in Section \ref{subsec:mayn}. We
    report how often the first-token of the answer changes ($\text{top}{-1}$)
    with the interventions $M(G)$ and $M(G^C)$. Statistically significant
    changes are marked in \textbf{bold} ($\chi^2$ test, $\alpha = 0.01$,
    $\alpha' = 2.2\mathrm{e}{-4}$).
    }
    \label{tab:graph-intervene-acc}
\end{table*}

\section{Appendix: Experimental Details}
\label{sec:appendix-experiments}

\begin{figure*}[t!]
    \centering
    \begin{subfigure}[t]{0.49\textwidth}
        \begin{center}
            \begin{tcolorbox}[width=\linewidth,colback=wrongcol,colframe=black, equal height group=B]
            \footnotesize{
            \textbf{Question: } Marin and his neighbor Nancy each eat 4 apples a day. How many apples do they eat in 30 days?\\
            \textbf{Ground Truth: } \textcolor{red}{150}  \textcolor{corgreen}{[240]}\\
            \textbf{Provided Reasoning: } In one day, Marin and Nancy eat \textcolor{red}{4 + 1} = <<4+1=5>>5 apples. In 30 days, they eat 30 * 5 = <<30*5=150>>150 apples.\\
            \textbf{Model Reasoning: } \\
              Apples eaten per person per day: $4$.  \\
              Number of people: $2$.  \\
              Total apples per day: $4 \times 2 = 8$.  \\
              Number of days: $30$.  \\
              Total apples in $30$ days: $8 \times 30 = 240$.
            }
            \end{tcolorbox}
        \end{center}
        \subcaption{\textbf{GSM8K}-Wrong ground truth label.}
    \end{subfigure}
    \begin{subfigure}[t]{0.49\textwidth}
        \begin{center}
            \begin{tcolorbox}[width=\linewidth,colback=wrongcol,colframe=black, equal height group=B]
            \footnotesize{
                \textbf{Question: } Gene is sewing a quilt out of old souvenir t-shirts. He has one shirt from each vacation he has been on. Every shirt is its own quilt block. Each row is made of blocks from a different year of vacations. He goes on four vacations a year and has been vacationing since he was 23 years old. He is now 34. How many quilt blocks does he have in total? \\
                \textbf{Ground Truth: } \textcolor{red}{44} \\
                \textbf{Model Answer: } \textcolor{red}{48} \\
                \textbf{Ambiguity: } Ages 23-34 can be 11 years (44) or 12 years (48) depending on the time of year. 
            }
            \end{tcolorbox}
        \end{center}
        \subcaption{\textbf{GSM8K} - Ambiguous question.}
    \end{subfigure}
    \begin{subfigure}[t]{0.49\textwidth}
        \begin{center}
            \begin{tcolorbox}[width=\linewidth,colback=boxcol,colframe=black, equal height group=C]
            \footnotesize{
                \textbf{Question: } A sequence $(a_n)$ is defined as follows:\\
                $a_{i + 1} = \frac{1}{1 - a_i}$ for $i \ge 1$. If $a_3 = a_1$,
                compute $(a_9)^9$.\\
                \textbf{Model Reasoning: } \\
                Let's think step by step.\\
                - Let $a_1=x$.\\
                - Define $f(z)=\frac{1}{1-z}$, so $a_{n+1}=f(a_n)$.\\
                - Then $a_2=f(x)=\frac{1}{1-x}$.\\
                - Compute $a_3=f(a_2)=f(\frac{1}{1-x})=\frac{1}{1-\frac{1}{1-x}}=\frac{1-x}{-x}=-\frac{1-x}{x}$.\\
                - Given $a_3=a_1$, we have $-\frac{1-x}{x}=x\Longrightarrow x^2-x+1=0$.\\
                - The roots are $x=\frac{1\pm i\sqrt{3}}{2}$, which satisfy $x^3=-1$ and $x^6=1$.\\
                - Since $f^2(x)=x$ (verified by the equation), the sequence is periodic with period $2$: $a_{2k+1}=x$, $a_{2k+2}=f(x)$.\\
                - Index $9$ is odd $\Rightarrow a_9=x$.\\
                - Using $x^6=1$, we have $x^9=x^{6}x^{3}=1\cdot(-1)=-1$.\\
                Answer: $\boxed{-1}$\\
                \textbf{Ground Truth: } -1

            }
            \end{tcolorbox}
        \end{center}
        \subcaption{\textbf{MATH500} - Example.}
    \end{subfigure}
    \begin{subfigure}[t]{0.49\textwidth}
        \begin{center}
            \begin{tcolorbox}[width=\linewidth,colback=boxcol,colframe=black, equal height group=C]
            \footnotesize{
                \textbf{Question: } Let $x$ , $y$ and $z$ all exceed $1$ and let
                $w$ be a positive number such that $\log_xw=24$,
                $\log_y w = 40$ and $\log_{xyz}w=12$. Find $\log_zw$.\\
                \textbf{Model Reasoning: } \\
                Let's think step by step.\\
                - $\log_x w = 24$ gives $w = x^{24}$.\\
                - $\log_y w = 40$ gives $w = y^{40}$.\\
                - $\log_{xyz} w = 12$ gives $w = (xyz)^{12}$.\\
                - Equating $x^{24}$ and $(xyz)^{12}$: $(xyz)^{12} = x^{24}$, so $xyz = x^{2}$.\\
                - Divide by $x$: $yz = x$.\\
                - From $w = x^{24} = y^{40}$, express $y$ in terms of $x$: $y = x^{24/40} = x^{3/5}$.\\
                - Let $\log_z w = L$, then $w = z^{L}$, so $z = x^{24/L}$.\\
                - Substitute $y$ and $z$ into $yz = x$: $x^{3/5} \cdot x^{24/L} = x$.\\
                - Combine exponents: $x^{3/5 + 24/L} = x^{1}$.\\
                - Exponents must match: $3/5 + 24/L = 1$.\\
                - Solve for $L$: $24/L = 1 - 3/5 = 2/5$, hence $L = 24 \cdot 5 / 2 = 60$.\\
                Answer: $\boxed{60}$\\
                \textbf{Ground Truth: } 60
            }
            \end{tcolorbox}
        \end{center}
        \subcaption{\textbf{AIME} - Example.}
    \end{subfigure}
    \caption{\textbf{Dataset samples.} \emph{(Top)} Examples of annotation
    errors in GSM8K. \emph{(Bottom)} Examples from MATH500 and AIME.}
    \label{fig:example_dataset}
\end{figure*}

\subsection{Data Curation - GSM8K}

$1000$ problems were chosen at random from the GSM8K test set for this study.
However, we observed inaccuracies in some of the provided solutions and
ground truth answers ($\sim2\%$), based on disagreement between OpenAI
\texttt{o3} responses and the ground truth. These broadly fall into two camps:
(i) wrong ground truth answer, and (ii) question ambiguity. $9$ ambiguous
questions were found, which were removed and $11$ wrong answers were found,
which were corrected. The final dataset contains $\mathbf{991}$ samples, of
which $\mathbf{983}$ were successfully parsed. $\mathbf{5}$ additional examples
were chosen at random from the training set to serve as in-context \emph{CoT}
demonstrations. An example problem from the dataset is in Figure
\ref{fig:example_annotation}, and examples of discovered annotation errors can
be found in Figure \ref{fig:example_dataset}.

\subsection{Data Curation - MATH500 \& AIME}

In these splits problems involving geometry are present, which pose a challenge
from the expression parsing perspective. As an example, consider the statement -
``\emph{Let the midpoint of line $\overline{AB}$ be $O$, from which...}''. It is
difficult to decipher the relationship between $O$ from algebraic parsing alone.
Further, some examples were found which refer to a diagram, but do not contain
the diagram, such as the following example:
\begin{tcolorbox}[width=\linewidth,colback=wrongcol,colframe=black] 
    \footnotesize{
        \textbf{MATH500 Question:} [\emph{Missing Diagram}]\\
        In the circle with center $Q$, radii $AQ$ and $BQ$ form a right angle.
        The two smaller regions are tangent semicircles,
        \textcolor{red}{\textbf{as shown}}. The radius of the circle with center
        $Q$ is 14 inches. What is the radius of the smaller semicircle? Express
        your answer as a common fraction.
    }
\end{tcolorbox}

Thus, for the purposes of the study we filtered out all geometric problems or
problems refering to diagrams with keyword based filters (e.g.,
``\texttt{diagram}'', ``\texttt{[asy]}'', ``\texttt{trapezoid}'',
``\texttt{quadrilateral}'', etc.). Following this filtering process, all
remaining ($389$) samples from MATH500, and $350$ random samples from AIME were 
used in the study. $5$ disjoint in-context samples were also chosen from each
split.

\subsection{Prompts}
\label{sec:appendix-prompts}

The system prompt for generating reasoning responses from OpenAI \texttt{o3} is
presented in Figure \ref{fig:prompt}.

\begin{figure*}[!t]
    \begin{center}
        \begin{tcolorbox}[width=\linewidth,colback=boxcol,colframe=black]
\# Identity\\\\
You are a helpful AI assistant that solves math problems. Answer the user's question after thinking step by step.\\\\
\# Instructions\\\\
* Do not produce extraneous text, stay on topic and only produce text relevant to the problem.\\
* When given a question think step by step.\\
* For all intermediate steps, show and explain all calculations.\\
* Write each thinking step on a different line starting with \verb!'  -'!.\\
* All mathematical expressions must be written in LaTeX notation, and enclosed inside '\$' signs.\\
* Do not write multi-line equation, if needed break them into multiple single line equations.\\
* Make all LaTeX inline to the best of your ability. Do not use multi-line equations, like align, array, etc., just write on different lines.\\
* Do not put text or sentence punctuation inside formulas, write them outside \$\$ signs. Avoid \verb!\text{}, \mbox{}!, etc.\\
* Finally, produce the answer in a latex box, such as \verb!$\boxed{53}$!.\\
* \textcolor{red}{\emph{The answer will always be a number}}, do not add units or anything else. Only the number, which is the final answer should be in \verb!$\boxed{}$!.\\
\\
\# Examples\\\\
<user>\\
Question: \textcolor{blue}{\textbf{Example in-context demonstration question.}}\\
</user>\\\\
<assistant>\\
Let's think step by step.\\
  - \textcolor{blue}{\textbf{Example in-context demonstration reasoning line 1.}}\\
  - \textcolor{blue}{\textbf{Example in-context demonstration reasoning line 2.}}\\
  - \textcolor{blue}{\textbf{...}}\\
Answer: \verb!$\boxed{! \textcolor{blue}{\textbf{Example answer.}}\verb!}$!\\
</assistant>\\
...
        \end{tcolorbox}
    \end{center}
    \caption{\textbf{System prompt for experiments.} We employed \emph{5-shot CoT}
    prompts, alongside general instructions for all experiments in our study.
    The examples were chosen at random and the reasoning demonstrations were
    created manually. There are minor variations in the prompt templates for the
    3 splits, e.g., the segment highlighted in red and italicized is specific to
    AIME and GSM8K, and is omitted for MATH500.
    }
    \label{fig:prompt}
\end{figure*}

\section{Appendix: Additional Examples}
\label{sec:appendix-path-examp}

Further examples of \textbf{R paths} from the GSM8K split of \textbf{KisMATH}
is presented in Figure \ref{fig:path_examples}.

\begin{figure*}[t!]
    \begin{center}
    \begin{subfigure}[t]{0.49\textwidth}

        \begin{tcolorbox}[width=\linewidth,colback=wrongcol,colframe=black, equal height group=aa] \footnotesize{
            \textbf{Question}: Larry cooked dumplings for a group of friends.  There are 8 males, including Larry, and \colorbox{qcolor}{6} females among them. Each male ate 1 more dumpling than each female. How many dumplings did Larry cook if each female ate 3 dumplings and there were no leftovers?\\
            \textbf{{Reasoning}}:
            Let's think step by step.  
            - Number of females: $\hllbox{6}$.  
            - Dumplings per female: $3$.  
            - Dumplings eaten by females: $\hllbox{6 \times 3 }= 18$.  
            - Number of males: $8$.  
            - Dumplings per male: $\hllbox{3 + 1 }= 4$.  
            - Dumplings eaten by males: $8 \times 4 =\hllbox{ 32}$.  
            - Total dumplings cooked: $18 + 32 =\hllbox{ 50}$.\\\textbf{Answer}: \colorbox{answercol}{50}
        }\end{tcolorbox}

        \begin{tcolorbox}[width=\linewidth,colback=wrongcol,colframe=black, equal height group=bb] \footnotesize{
            \textbf{Question}: Zeke's baseball team has 7 more players than Carlton's. If Carlton's team has \colorbox{qcolor}{13} players, how many players are there in both teams combined?\\
            \textbf{{Reasoning}}:
            Let's think step by step.  
            - Carlton’s team has $13$ players.  
            - Zeke’s team has $13 + 7 =\hllbox{ 20}$ players.  
            - Total players on both teams: $13 + 20 =\hllbox{ 33}$.\\\textbf{Answer}: \colorbox{answercol}{33}
        }\end{tcolorbox}

        \begin{tcolorbox}[width=\linewidth,colback=wrongcol,colframe=black, equal height group=cc] \footnotesize{
            \textbf{Question}: Jon runs a triathlon.  It takes him \colorbox{qcolor}{40} minutes for the swim, an hour and 20 minutes for the bike ride and 50 minutes for the run.  Compared to Jon, James finishes the swim 10\% faster but takes 5 minutes longer on the bike.  If Jon won by 10 minutes, how long did it take James to do the run?\\
            \textbf{{Reasoning}}:
            Let's think step by step.  
            - Jon’s times: swim $\hllbox{40}$, bike $1\text{ hr }20\text{ min}=80$, run $50$.  
            - Jon’s total: $\hllbox{40+80+50}=170$.  
            - James’s swim: $\hllbox{40(1-0.10)}=36$.  
            - James’s bike: $80+5=85$.  
            - Let $R$ be James’s run time.  
            - James’s total: $36+85+R=\hllbox{121+R}$.  
            - Jon wins by $10$ minutes: $\hllbox{121+R }= 170+10 = 180$.  
            - Solve for $\hllbox{R}$: $R = 180-121 =\hllbox{ 59}$.\\\textbf{Answer}: \colorbox{answercol}{59}
        }\end{tcolorbox}

        \begin{tcolorbox}[width=\linewidth,colback=wrongcol,colframe=black, equal height group=D] \footnotesize{
            \textbf{Question}: Adrianne is collecting different kinds of beads for making bracelets. Her mother gave her \colorbox{qcolor}{20} metallic beads. Her sister gave her ten more beads than her mother, and her friend gave her twice as many as her mother gave.  How many beads did Adrianne have altogether?\\
            \textbf{{Reasoning}}:
            Let's think step by step.  
            - Beads from mother: $20$.  
            - Beads from sister: $\hllbox{20 + 10 }= 30$.  
            - Beads from friend: $2 \times 20 =\hllbox{ 40}$.  
            - Total beads: $20 + 30 + 40 =\hllbox{ 90}$.\\\textbf{Answer}: \colorbox{answercol}{90}
        }\end{tcolorbox}

    \end{subfigure}
    \begin{subfigure}[t]{0.49\textwidth}

        \begin{tcolorbox}[width=\linewidth,colback=wrongcol,colframe=black, equal height group=aa] \footnotesize{
            \textbf{Question}: At the end of the school year, Kate asked her teacher if she could have the 3 boxes of \colorbox{qcolor}{64} crayons since they were all worn down to small pieces.  The teacher agreed.  Kate took the crayons home and decided to make crayon muffins by melting 8 small pieces of crayons together in a muffin tin that she placed in an oven for a few minutes. If she can sell her muffin crayons for \$1.50 each, how much money can Kate make?\\
            \textbf{{Reasoning}}:
            - Total crayon pieces: $\hllbox{3 \times 64 }= 192$
            - Muffins that can be made: $192 \div 8 =\hllbox{ 24}$
            - Revenue per muffin: USD 1.50
            - Total revenue: $24 \times 1.50 =\hllbox{ 36}$\\\textbf{Answer}: \colorbox{answercol}{36}
        }\end{tcolorbox}

        \begin{tcolorbox}[width=\linewidth,colback=wrongcol,colframe=black, equal height group=bb] \footnotesize{
            \textbf{Question}: Sophia and Rose went together to the market to buy onions and potatoes. Rose bought 4 times the number of onions and potatoes Sophia bought. If Rose bought \colorbox{qcolor}{12} onions and 4 potatoes, how many onions and potatoes in total did Sophia buy at the market?\\
            \textbf{{Reasoning}}:
            - Rose’s total purchase: $\hllbox{12 + 4 }= 16$ items.  
            - Rose bought $\hllbox{4}$ times as many items as Sophia, so Sophia’s total is $\dfrac{16}{4} = \hllbox{4}$ items.\\\textbf{Answer}: \colorbox{answercol}{4}
        }\end{tcolorbox}

        \begin{tcolorbox}[width=\linewidth,colback=wrongcol,colframe=black, equal height group=cc] \footnotesize{
            \textbf{Question}: George, a grade six teacher, ordered 600 burritos for the sixth-grade class picnic. If there were \colorbox{qcolor}{50} students at the picnic, and each student was given ten burritos, with Mr. George eating 20 of them, calculate the total number of leftover burritos from the picnic?\\
            \textbf{{Reasoning}}:
            - Total burritos ordered: $600$.
            - Burritos eaten by students: $\hllbox{50 \times 10 }= 500$.
            - Burritos eaten by Mr.George: $20$.
            - Total burritos eaten: $500 + 20 =\hllbox{ 520}$.
            - Leftover burritos: $600 - 520 =\hllbox{ 80}$.\\\textbf{Answer}: \colorbox{answercol}{80}
        }\end{tcolorbox}

        \begin{tcolorbox}[width=\linewidth,colback=wrongcol,colframe=black, equal height group=D] \footnotesize{
                \textbf{Question}: John and Jack have 30 minutes to walk to school together. It takes them 6 minutes to get to the corner where the library is. It takes them another \colorbox{qcolor}{13} minutes to get to the fire station. How much longer do they have to get to school without being late?\\
            \textbf{{Reasoning}}:
            Let's think step by step.  
            - Total time available: $30$ minutes.  
            - Time to the library corner: $6$ minutes.  
            - Additional time to the fire station: $\hllbox{13}$ minutes.  
            - Time already spent: $6 + 13 =\hllbox{ 19}$ minutes.  
            - Remaining time: $30 - 19 =\hllbox{ 11}$ minutes.\\\textbf{Answer}: \colorbox{answercol}{11}
        }\end{tcolorbox}
    \end{subfigure}
    \end{center}
    \caption{Examples of \textbf{R paths} from the GSM8K split of the \textbf{KisMATH} dataset.}
    \label{fig:path_examples}
\end{figure*}

\end{document}